\documentclass{article}

\usepackage[utf8]{inputenc}
\usepackage{url}
\usepackage{bigstrut}
\usepackage{microtype}
\usepackage{graphicx}
\usepackage{amsfonts}
\usepackage{nicefrac}
\usepackage[breaklinks,pagebackref=false,breaklinks=true,colorlinks,bookmarks=false]{hyperref}
\usepackage{subfigure}
\usepackage{xspace}
\usepackage{booktabs} 
\usepackage{pdfpages}
\usepackage{subfigure}
\usepackage[notransparent]{svg}
\usepackage{svg-extract}
\usepackage{listings}
\definecolor{codegreen}{rgb}{0,0.6,0}
\definecolor{codegray}{rgb}{0.5,0.5,0.5}
\definecolor{codepurple}{rgb}{0.58,0,0.82}
\definecolor{backcolour}{rgb}{0.95,0.95,0.92}
\usepackage[normalem]{ulem}
\useunder{\uline}{\ul}{}
\usepackage{colortbl}
\usepackage{color,xcolor,xspace}
\usepackage{multirow}
\lstdefinestyle{mystyle}{
    backgroundcolor=\color{backcolour},   
    commentstyle=\color{codegreen},
    keywordstyle=\color{magenta},
    numberstyle=\tiny\color{codegray},
    stringstyle=\color{codepurple},
    basicstyle=\ttfamily\footnotesize,
    breakatwhitespace=false,
    breaklines=true,                 
    captionpos=b,                    
    keepspaces=true,                 
    showspaces=false,                
    showstringspaces=false,
    showtabs=false,                  
    tabsize=2
}

\usepackage[accepted]{icml2025}
\usepackage{amsmath}
\usepackage{amssymb}
\usepackage{mathtools}
\usepackage{amsthm}

\usepackage[capitalize,noabbrev]{cleveref}

\theoremstyle{plain}

\theoremstyle{definition}

\theoremstyle{remark}

\usepackage[textsize=tiny]{todonotes}

\definecolor{light-gray}{gray}{0.9}
\definecolor{light-gray0}{gray}{0.7}
\newcommand\best[1]{\textcolor{red}{\textbf{#1}}}
\newcommand\secbest[1]{\textcolor{blue}{\ul #1}}
\newcommand{\Ours}{\text{Captain Agent}\xspace}
\newcommand*\circled[1]{\tikz[baseline=(char.base)]{
            \node[shape=circle,draw,inner sep=2pt] (char) {#1};}}

\icmltitlerunning{Adaptive In-conversation Team Building for Language Model Agents}

\begin{document}

\twocolumn[
\icmltitle{Adaptive In-conversation Team Building for Language Model Agents}



\icmlsetsymbol{equal}{*}

\begin{icmlauthorlist}
\icmlauthor{Linxin Song}{equal,usc}
\icmlauthor{Jiale Liu}{equal,psu}
\icmlauthor{Jieyu Zhang}{uw}
\icmlauthor{Shaokun Zhang}{psu}
\icmlauthor{Ao Luo}{wsd}
\icmlauthor{Shijian Wang}{}\\
\icmlauthor{Qingyun Wu}{psu}
\icmlauthor{Chi Wang}{deepmind}
\end{icmlauthorlist}

\icmlaffiliation{usc}{University of Southern California}
\icmlaffiliation{psu}{Pennsylvania State University}
\icmlaffiliation{uw}{University of Washington}
\icmlaffiliation{wsd}{Waseda University}
\icmlaffiliation{deepmind}{Google Deepmind}

\icmlcorrespondingauthor{Chi Wang}{chi@autogen.team}

\icmlkeywords{LLM, Agent, Multi-agent}

\vskip 0.3in
]



\printAffiliationsAndNotice{\icmlEqualContribution} 

\begin{abstract}
Leveraging multiple large language model (LLM) agents has shown to be a promising approach for tackling complex tasks, while the effective design of multiple agents for a particular application remains an art. It is thus intriguing to answer a critical question: \textit{Given a task, how can we build a team of LLM agents to solve it effectively?} Our new adaptive team-building paradigm offers a flexible solution, realized through a novel agent design named \emph{\Ours}. It dynamically forms and manages teams for each step of a task-solving process, utilizing nested group conversations and reflection to ensure diverse expertise and prevent stereotypical outputs, allowing for a flexible yet structured approach to problem-solving. A comprehensive evaluation across six real-world scenarios demonstrates that \Ours significantly outperforms existing multi-agent methods with 21.94\% improvement in average accuracy, providing outstanding performance without requiring task-specific prompt engineering. Our exploration of different backbone LLM and cost analysis further shows that \Ours can improve the conversation quality of weak LLM and achieve competitive performance with extremely low cost, which illuminates the application of multi-agent systems.
\end{abstract}

\section{Introduction}
The success of large language model (LLM) agents~\citep{yao2022react, yang2023auto, furuta2024multimodal, yang2024sweagent, datainterpreter} with its outstanding in-context learning~\citep{dong2022survey,brown2020language, yang2023iterative,dai2023can,li2023transformers}, planning~\citep{sun2024adaplanner,xie2024travelplanner,Liu2023LLMPEL,Valmeekam2022PlanBenchAE,Wei2022ChainOT,Yuan2023DistillingSK,Zheng2024GPT4VisionIA}, tool-using~\citep{qin2023tool,qin2023toolllm,schick2024toolformer,cai2023large,yuan2023craft,paranjape2023art,zhang2024training, huang2023metatool,ma2024m}, and conversation~\citep{fernandes2023bridging,wang2023mint,yang2024intercode} capabilities allow us to relate human's team building and collaboration abilities to the multiple language model agents (multi-agent) system~\citep{wang2023survey, xi2023rise, wu2023autogen, meta-prompting, hong2023metagpt, zhang2024training, zhang2023ecoassistant, valmeekam2023planning, wang2024tdag,saha2023branch,liang2023encouraging,du2023improving,chen2024agentverse}. 
Humans have developed abilities that enable us to form teams and effectively solve problems. These abilities are rooted in communication, social cognition, problem-solving and decision-making, social learning and imitation, and shared intentionality~\citep{elimari2020network,confer2010evolutionary}. 
The interplay of the above abilities allows people to organize different teams for problems to ensure that tasks are completed successfully, which brings us to a critical question in a multi-agent system:

{\textit{Given a task, how can we build a team of LLM agents to solve it effectively?}}

A straightforward paradigm would be to build a static agent team beforehand based on the task instruction and let them solve the task collaboratively~\citep{autoagents, wu2023autogen}. However, \emph{static build} necessitates maintaining a team with all the required expertise for the whole task cycle. As the complexity of the task increases, the total number of team members may grow significantly. Always proceeding with such a large team makes it challenging to manage the team members effectively and efficiently. Furthermore, static teams may lack the adaptability to respond to dynamic changes in task requirements or unforeseen challenges.
Imagine a prehistoric human tribe: was everyone involved in every task? The answer is unlikely affirmative. Those responsible for hunting may not participate in medical care and those responsible for cooking may not involve themselves in management. The major task, survival, was ensured by each individual group sticking to their roles and subtasks. In fact, when human organizations handle a complex task, we tend to form multiple teams for each subtask at different stages of the task-solving procedure, which still guarantees a diverse set of expertise is leveraged demanded by the task complexity~\citep{mao2016experimental}.

Inspired by how humans assemble teams for a complex task, we introduce a new multi-agent team-building paradigm: \emph{adaptive build}. This paradigm facilitates the flexible assembly of agents with specific skills and knowledge as demands evolve in the process of task-solving.
To realize this paradigm, we propose a new adaptive builder agent, \Ours, to build, manage, and maintain agent teams for each problem-solving step in the conversation. \Ours has two core components: (1) adaptive multi-agent team building and (2) nested group conversation and reflection. 
\Ours will communicate with a User Proxy, who can provide the general task instructions at the beginning.
When assigned a task, \Ours begins by formulating a strategic plan. This plan involves a cyclical process that continues until the task is successfully completed. In the first phase of the cycle, \Ours identifies a specific subtask, outlines the necessary roles, and assembles a team of agents equipped with the appropriate tools. In the subsequent phase, this team engages in a dialogue with a versatile tool to address the subtask. Upon completion, a reflector LLM reviews the process and provides \Ours with a detailed reflection report. Based on this feedback, \Ours either adjusts the team composition or the subtask instructions and repeats the cycle or concludes the task and presents the final outcomes.

We evaluate state-of-the-art multi-agent approaches for complex task solving and our adaptive build approach with \Ours on six real-world scenarios, including many mathematics problem-solving~\citep{math}, data analysis~\citep{dabench}, programming~\citep{humaneval}, scientific problem-solving~\citep{scibench} (Physics and Chemistry), and information retrieval~\citep{gaia}.
Our experimental results demonstrated the outstanding ability of \Ours in various scenarios without prompt engineering. \Ours achieves distinguishing results compared to other single and multi-agent methods and frameworks when using the same prompt for each task, with an average of 21.94\% improvement on average accuracy. Ablation studies on static and adaptive building paradigms show that the adaptive team outperforms the static team in four of five scenarios (and matches in one scenario), exhibiting the superiority of the adaptive build paradigm across different scenarios. 
We also demonstrated that handcraft agents and handcraft tools contribute equally to the final results. 
We further explore the influence of different backbone LLM for both \Ours and nested group chat members or only for nested group chat members. We observe that: (1) \Ours with a strong backbone can improve the quality of nested group chat in which the members equipped with weak backbone, and (2) a small model with distinguishable instruction following ability can achieve outstanding performance with low cost.

\begin{figure}[t!]
    \centering
    \includegraphics[width=\linewidth]{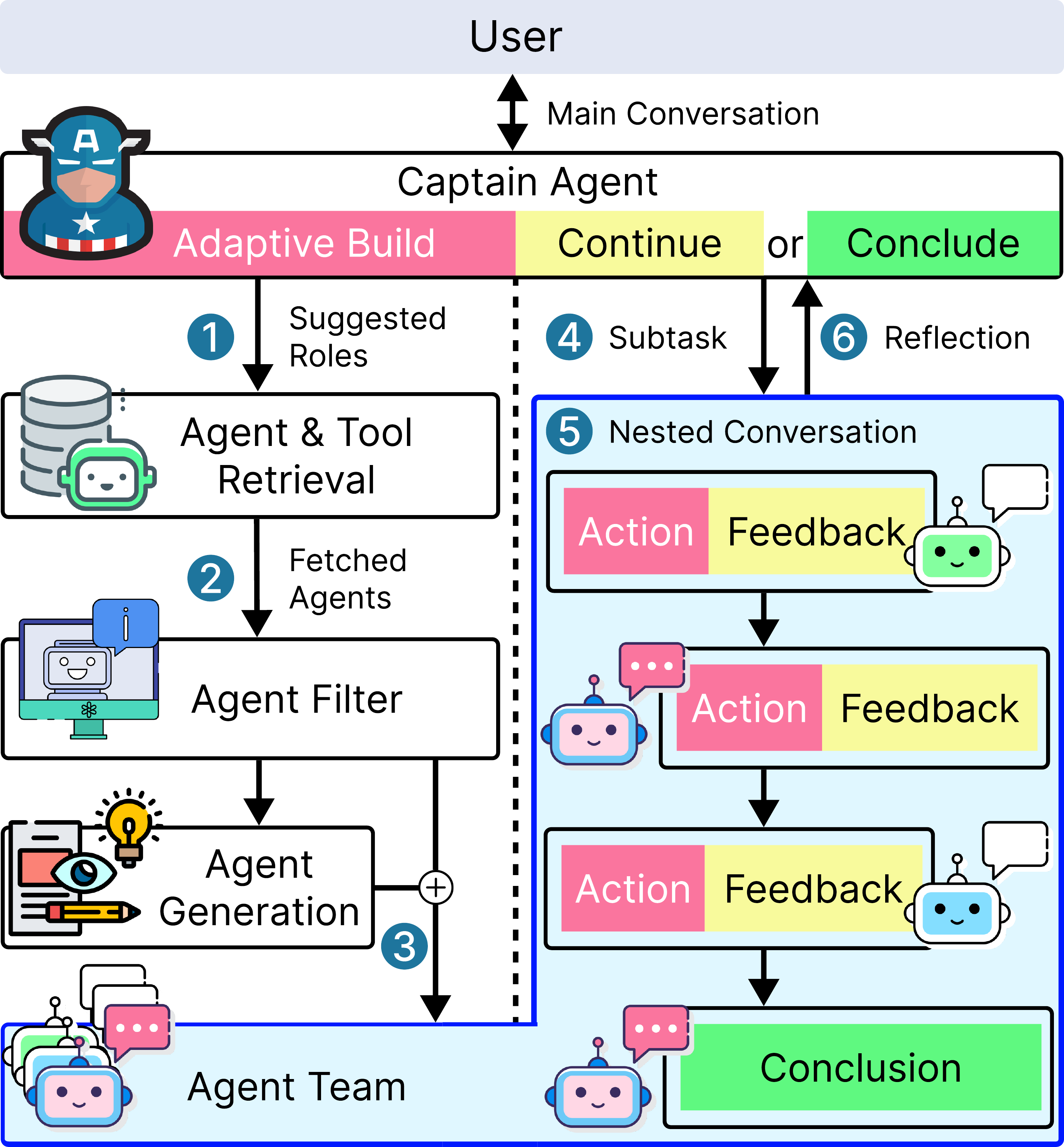}
    \caption{\label{fig:overall}The overall workflow of \Ours. \Ours can build an agent team apart from the main conversation and make next-step decisions according to the nested conversation results. We highlight the order of reading in the figure and mark them in the following sections.}
\end{figure}
\section{Captain Agent}



The overall workflow of \Ours is illustrated in Figure~\ref{fig:overall}. We first establish a main conversation between the user and \Ours. When receiving a user task, \Ours will: (\textbf{Step 1}) first decompose the task into a list of subtasks and suggest several roles for this subtask. A team of agents will be created accordingly. Each agent will be equipped with several predefined tools according to the abilities defined by \Ours (Section~\ref{sec:team-building}); (\textbf{Step 2}) the agent team will attempt to solve a subtask via a nested conversation. Once the nested conversation reaches the terminal condition, the conversation history will be summarized and analyzed by a reflector LLM, and the final result will be feedback to \Ours (Section~\ref{sec:conversation}).
These two steps will repeat certain rounds in the main conversation until reaching the max round limitation or terminal condition.

\subsection{Adaptive Multi-agent Team Building}
\label{sec:team-building}
After identifying a subtask in Step 1 following a corresponding prompt, \Ours will list several roles for the subtask. These roles will then pass into a retrieval, selection, and generation process guided by Retrieval-Augmented Generation (RAG)~\citep{lewis2020retrieval, gao2023retrieval, ram2023context}. Created agents will be equipped with a well-designed profile (system message\footnote{System message refers to a profile that defines an agent's persona and task-specific instructions.}) and high-quality tools.

\noindent\circled{1}\textbf{Agent and tool retrieval.}
\Ours will suggest $n$ required roles $R^{(j)}=\{r_{ij}| i\in {1,\cdots, n}\}$ for the incoming user task at round $j$. $r_{ij}$ includes a short sentence that can describe itself. We then retrieve top-$k_1$ agents and top-$k_2$ tools for $r_{ij}$ according to the embedding similarity between $r_{ij}$ and the agent/tool description in the library, as follows:
\begin{align}
    &C_{\text{agent}}^{(i,j)} = \text{top-}k_1 \ \text{CosineSimilarity}\left(f(r_{ij}), f(A_{\text{lib}})\right), \notag\\
    &C_{\text{tool}}^{(i, j)} = \text{top-}k_2 \ \text{CosineSimilarity}\left(f(r_{ij}), f(T_{\text{lib}})\right), \notag
\end{align}
where $k_1$ and $k_2$ are the amount of retrieved agents and tools from agent library $A_{\text{lib}}$ and tool library $T_{\text{lib}}$, respectively. $f(\cdot)\in\mathbb{R}^{m}$ denotes the extraction of the embeddings. We use $C$ to denote the word "candidate."
We then equip $C_{\text{agent}}$ with corresponding $C_{\text{tool}}$.

\noindent\circled{2}\textbf{Agent filtering.}
Finding an exactly matched agent from our agent library for every real-world task is nearly impossible. Therefore, we prompt an LLM-based filter to select the top-$1$ suitable agent for $r_{ij}$ from $C^{(i,j)}_{\text{agent}}$, simulating the human selection process from massive retrieved results (e.g., searching on the internet). We provide candidate agents $C_{\text{agent}}$, suggested role $r_{ij}$, and the user task as input to the filter LLM. The filter LLM is required to return the \textit{most suitable one} agent from $C_{\text{agent}}$. An abstention mechanism is adopted to this process, i.e., the filter LLM can return \textit{None} to prevent irrelevant agents from being selected as team members. This often happens when $|A_{\text{lib}}|$ is small. We then go through an agent generation process for those $r_{ij}$ with no suitable agents.

\noindent\circled{3}\textbf{Agent generation.}
We design a step-by-step and templated agent generation process for $r_{ij}$ with no suitable agent selected at the previous step, motivated by \citet{wei2022chain, shinn2024reflexion}. Generally, it includes three steps: (1) generate a basic profile, including name, persona, and required skills as the system message; (2) combine the basic system message with predefined general task-solving and environment interaction instructions, for example, group chat information; (3) retrieved suitable tools from $T_{\text{lib}}$ according to the generated system message and bind the new agent with retrieved tools. We then update the newly generated agent into $A_{\text{lib}}$ for future use.


\textbf{Team memory.}
\Ours will cache the built team into its conversation-wise temporary memory, which only includes basic details for each member of the team and no previous conversation history. \Ours can prompt the same team with new instructions to seek an alternate result without going through the team-building process, which helps reduce the total cost.


\subsection{Nested Conversation and Reflection}
\label{sec:conversation}
Once the team-building process is completed, a new conversation will be established between the created agent team members apart from the main conversation, which we call "nested conversation." \Ours will provide a subtask to the nested conversation. At the end of the nested conversation, an LLM-based reflector will analyze the whole conversation history. \Ours next-step decision (continue or conclude) will based on the reflector's report.

\noindent\circled{4,5}\textbf{Nested conversation.} 
In the nested conversation, an LLM-based conversation manager will adaptively select the next speaker from the agent team members based on the conversation history.
Team members will share the same conversation history in the nested group chat, but they cannot access the history from the main conversation.
Conversation history consists of the previous agent's action and corresponding feedback.
Each agent's action can be natural language instruction and/or a Python code block.
Specifically, we use raw Python function as our tool format, which allows agents to create new tools with their assigned tools $C_{\text{tool}}^{(i, j)}$ by incorporating $C_{\text{tool}}^{(i, j)}$ into a self-written Python code block, which helps extend agent's action set creatively.

\noindent\circled{6}\textbf{Conversation reflection.}
A conversation can easily escalate into a conflict once an agent starts a dispute with another, where they will never make a consensus and continue until reaching the max-round. This often happens when \Ours provides an ambiguous instruction or the subtask is not clear enough. Therefore, we design a reflection mechanism for nested conversation. An LLM-based reflector will summarize the conversation history and check all disputes inside the conversation, then provide feedback to the \Ours with a reflection report. \Ours will make further decisions based on this report. All reflection reports will be recorded in the main conversation.

\begin{table*}[htbp]
\caption{\label{tab:dataset}Scenarios and the corresponding datasets. 
We perform the main comparison experiments on the whole dataset except MATH, which sampled a subset according to the type distribution.
}
\resizebox{\textwidth}{!}{%
\begin{tabular}{lccl}
\hline
\rowcolor{light-gray0} \textbf{Scenario} &
  \textbf{Dataset} &
  \textbf{Size} &
  \textbf{Sample} \bigstrut\\ \hline
Mathematics problems &
  MATH~\citep{hendrycksmath2021} &
  196 &
  \renewcommand\arraystretch{1} \begin{tabular}[c]{@{}l@{}}If $\frac{3x^2-4x+1}{x-1}=m$, and $x$ can be any real number except $1$, \\ what real values can $m$ NOT have?\end{tabular} \bigstrut\\
\rowcolor{light-gray} Programming &
  HumanEval~\citep{chen2021codex} &
  164 &
  \renewcommand\arraystretch{1} \begin{tabular}[c]{@{}l@{}}def truncate\_number(number: float) -\textgreater float:\\\quad""" Given a positive floating point number, it can be decomposed into\\\quad and integer part (largest integer smaller than given number) and decimals\\\quad (leftover part always smaller than 1).\\\quad {[}Omitted{]}\\\quad"""\end{tabular} \bigstrut\\
Data Analysis &
  DABench~\citep{hu2024infiagentdabench} &
  257 &
  \renewcommand\arraystretch{1} \begin{tabular}[c]{@{}l@{}}Generate a new feature called "FamilySize" by summing the "SibSp" \\ and "Parch" columns. Then, calculate the Pearson correlation coefficient (r) \\ between the "FamilySize" and "Fare" columns.\end{tabular} \bigstrut\\
\rowcolor{light-gray} information retrieval &
  GAIA~\citep{mialon2023gaia} &
  165 &
  \renewcommand\arraystretch{1} \begin{tabular}[c]{@{}l@{}}On the BBC Earth YouTube video of the Top 5 Silliest Animal Moments, \\ what species of bird is featured?\end{tabular} \\
(Scientific) Chemistry &
  SciBench~\citep{scibench} &
  41 &
  \renewcommand\arraystretch{1} \begin{tabular}[c]{@{}l@{}}Calculate the pressure in kilopascals exerted by $1.25 \mathrm{~g}$  of nitrogen gas\\ in a flask of volume $250 \mathrm{~cm}^3$ at $20^{\circ} \mathrm{C}$.\end{tabular} \\
\rowcolor{light-gray} (Scientific) Physics &
  SciBench~\citep{scibench} &
  34 &
  \renewcommand\arraystretch{1} \begin{tabular}[c]{@{}l@{}}If the coefficient of static friction between the block and plane in the \\ previous example is $\mu_s=0.4$, at what angle $\theta$ will the block starts sliding \\ if it is initially at rest?\end{tabular} \bigstrut\\ \hline
\end{tabular}
}
\end{table*}
\subsection{Benefits over Static Team}
A static team may limit the team's ability coverage. Although building a large number of agents with comprehensive persona or skill sets can address the limitation in ability coverage, it is challenging for LLMs to handle a long context that introduces all the participant members. Unexpectedly long contexts will primarily reduce the quality of the conversation. Meanwhile, agents with redundant functionality will also be involved in the task-solving process. In contrast, \Ours can adaptively select and build more optimized agent teams for the current task, reducing the prompting load for LLMs and redundant output from irrelevant agents without sacrificing the diversity in the team. 

\section{Evaluation}
\subsection{Experimental Setup}
\label{sec:exp-setup}
\textbf{Scenarios and datasets.} We choose six real-world scenarios, including mathematics, programming, data analysis, information retrieval, and science problem-solving. Each scenario was chosen for its unique ability to demonstrate specific capabilities and performance metrics of the agent systems. This ensures a holistic assessment of \Ours against the baselines across various critical dimensions of computational and cognitive skills. We bind each scenario with a challenging open-source dataset, as shown in Table \ref{tab:dataset}. Due to cost limitations, we sample a subset of MATH according to its original distribution of each question type.

\textbf{Compared methods and implementation.}
For mathematics problems, programming, data analysis, and scientific scenarios, we investigate the performance of \Ours and four different methods, including Vanilla LLM (prompt an LLM once for an answer), AutoAgents~\citep{autoagents}, Meta-prompting~\citep{meta-prompting}, AgentVerse~\citep{chen2024agentverse}, DyLAN~\citep{liu2023dynamic}, and a two-agent system (a system involving an Assistant agent with an Executor agent) realized with AutoGen~\citep{wu2023autogen}. Specifically, we implement AutoAgents with AutoGen as the official implementation is unstable and unsuitable for large-scale experiments. For meta-prompting, we improve its code execution ability by reproducing a real-time Python interpreter. All these methods will be prompted with the same task-specific instruction (refer to Appendix~\ref{apdx:instructions}). 

For information retrieval scenarios, we compare \Ours with the top-5 baselines (with reference) reported to the GAIA validation leaderboard, which includes AutoGen: GAIA\_Orchestrator (a specific three-agent setting organized by an Orchestrator agent designed for GAIA)~\citep{orchestrator}, FRIDAY~\citep{friday}, Warm-up Act\footnote{Warm-up Act has no open-sourced implementation.}, GAIA official implementation, and AutoGPT~\citep{yang2023auto}.

For \Ours, we adopt \texttt{all-mpnet-base-v2} to calculate the sentence embedding for agent and tool retrieval. Our environment will communicate with \Ours by providing the feedback of code execution, tool calling (adaptive build), nested conversation reflection results, and a default reply: \textit{I'm a proxy, and I can only execute your code and tool or end the conversation. If you think the problem is solved, please reply to me only with 'TERMINATE.'}

\begin{table*}[ht!]
\caption{\label{tab:main-res}Comparison results on different real-world scenarios. We record each scenario's accuracy for each baseline and \Ours, and mark the best results in \textbf{bold}. We adopt \texttt{gpt-4-0125-preview} as the backbone LLM model \textbf{for all results.}}
\resizebox{\linewidth}{!}{
\renewcommand\arraystretch{1.3}
\begin{tabular}{lcccccc}
\hline \rowcolor{light-gray0} \textbf{Method}         & \textbf{Mathematics} & \textbf{Programming} & \textbf{Data Analysis} & \textbf{(Sci) Chemistry} & \textbf{(Sci) Physics} & \textbf{Avg.} \\ \hline
Vanilla LLM    & 51.53      & 84.76            & 6.61          & 39.02           & 31.25    &  40.98     \\
Meta-prompting & 68.88      & 19.51            & 39.69         & 41.46           & 43.75    &  43.47   \\
AutoAgents     & 56.12      & 84.76            & 57.98         & 60.98           & 50.00    &  63.58    \\
DyLAN     & 62.24      & 90.24            & -             & 45.45           & 51.16    &  -        \\ 
AgentVerse     & 69.38      & 42.68            & -             & 42.42           & 37.21    &  -        \\ 
AutoGen: Assistant + Executor        & 74.49      & 93.90            & 82.88         & 60.98           & 43.75   & 79.89      \\
\rowcolor{light-gray} \Ours  & \textbf{77.55}      &  \textbf{96.95}           & \textbf{88.32}         & \textbf{65.85}           & \textbf{53.12} & \textbf{84.25} \\\hline
\end{tabular}
}
\end{table*}

\begin{table*}[t!]
\caption{\label{tab:main-gaia}Comparison results on information retrieval scenario (GAIA validation). We report the accuracy at each level and the average accuracy over three levels and mark the best results in \textbf{bold}. \Ours achieves the best with minimal prompt engineering.}
\centering
\resizebox{0.7\linewidth}{!}{%
\renewcommand\arraystretch{1.3}
\begin{tabular}{lcccc}
\hline \rowcolor{light-gray0}
\textbf{Method}                         & \textbf{Level 1} & \textbf{Level 2} & \textbf{Level 3} & \textbf{Avg.}  \\ \hline
AutoGPT (\texttt{gpt-4}) & 13.21 & 0.00 & 3.85 & 4.85 \\
\texttt{gpt-4-turbo} & 20.75 & 5.81 & 0.00 & 9.70\\
\texttt{gpt-4-turbo} + manually selected plugins & 30.30 & 9.70 & 0.00 & 14.6 \\
Warm-up Act (\texttt{gpt-4-turbo})                    & 35.19   & 15.12   & 0       & 17.58 \\
FRIDAY (\texttt{gpt-4-turbo})               & 45.28   & 34.88   & \textbf{11.54}   & 34.55 \\
AutoGen: GAIA\_Orchestrator (\texttt{gpt-4-turbo})         & 54.72   & 38.31   & \textbf{11.54}   & 39.39 \\
\rowcolor{light-gray}\Ours (\texttt{gpt-4-turbo})     & \textbf{56.60}   & \textbf{39.53}   & \textbf{11.54}   & \textbf{40.60} \\ \hline
\end{tabular}%
}
\end{table*}

\textbf{Agent and tool library.} We initialize our agent library based on a small subset of problem instances from each dataset (described in Section~\ref{sec:abla}, about 20 questions for each dataset) in Table~\ref{tab:dataset}. 
Specifically, we run \Ours on the subset and iteratively update the library by adding the generated agents and keeping our agent library unchanged during the main experiment.
Our agent library also supports all hand-crafted agents (of the \texttt{ConversableAgent} class) archived in AutoGen (details in Appendix~\ref{apdx:agent-lib}). 
All these agents follow the ConversableAgent interface to converse with each other. 
Our tool library consists of a suite of callable Python functions intended for freeform coding. The agents can freely import functions from the tool library and write free-form code to integrate the outputs to handle sophisticated tasks (see also Appendix~\ref{apdx:case} and \ref{apdx:tool-lib}). 
The library contains three main categories of tools: math, data analysis, and information retrieval. For each category, we summarize the patterns of the corresponding dataset and manually craft a set of functions that suit the tasks.

\subsection{Evaluation Protocol} 
For mathematics, data analysis, and science scenarios, we report the accuracy of each method by comparing the final result from each method and ground truth. To ensure fairness in evaluation, we transform different result formats into a uniform format, preventing the correct answer from being judged incorrect due to format mismatches. For programming scenarios, we run the code provided from each method and output a unique token if the code successfully passes all tests. We then count the success token and calculate the accuracy for each method.

\subsection{Main Results}
\label{sec:main}

Table~\ref{tab:main-res} and \ref{tab:main-gaia} report the comparison results between \Ours and eight different baselines on six real-world scenarios. Baseline results on information retrieval are extracted directly from the GAIA leaderboard.

\textbf{Findings 1: Diverse agents can help trigger accurate expertise output for problem-solving.}
By comparing the results from \Ours, AutoAgents, and AutoGen Assistant + Executor, we observe that \Ours and AutoAgents averagely outperform AutoGen Assistant + Executor on (Sci) Chemistry and (Sci) Physics scenarios. These scenarios required expertise knowledge, which the AutoGen Assistant with a fixed system message is hard to complete. \Ours and AutoAgents can create diverse experts by assigning different domain-specific system messages to agents, which helps better trigger the intrinsic knowledge inside an LLM to provide an accurate answer.  \Ours outperforms AutoAgents in all the scenarios because \Ours can provide a high-level plan and solve each step with adaptive instructions and an agent team.

\textbf{Findings 2: Adaptive team-building boosts performance with no task preference.} It is obvious that \Ours achieves outstanding results over all scenarios, indicating that \Ours is free from task preference. Incorporating different agents into the team at a proper time gives \Ours the ability to solve difficult tasks like science and information retrieval problems step-by-step. 
We observe that Meta-prompting fails in science scenarios due to the inability to decompose science problems into the fine-grain subtasks that one agent can solve. 
\Ours with the agent-team building paradigm neither requires a task that can be decomposed into a subtask that can only be solved by an agent nor requires all agents to be involved in the conversation. 
Compared with other agent-team building methods such as AutoAgents, DyLAN, and AgentVerse, \Ours can not only 
We further discuss the static and adaptive teams in Section~\ref{sec:abla-team}.

\subsection{Analysis and Ablation Studies}
\label{sec:abla}
In this section, we dive into the difference between static and adaptive team-building, the influence of agent and tool libraries, and the possibility of working with open-weight models.
We perform ablation studies on a subset from Table~\ref{tab:dataset}. Specifically, we choose 17 problems from MATH and 25 problems from HumanEval according to the AutoGenBench~\citep{autogenbench}, in which the problems are randomly selected from GPT-4 failure set. For DABench, we randomly selected 25 problems, and for SciBench, we randomly selected 19 problems for chemistry and physics according to the number of textbooks. The evaluation protocol is the same as in Section~\ref{sec:main}.

\subsubsection{Static vs. adaptive team-building}
\label{sec:abla-team}
\begin{figure}[t!]
    \centering
    \includegraphics[width=\linewidth]{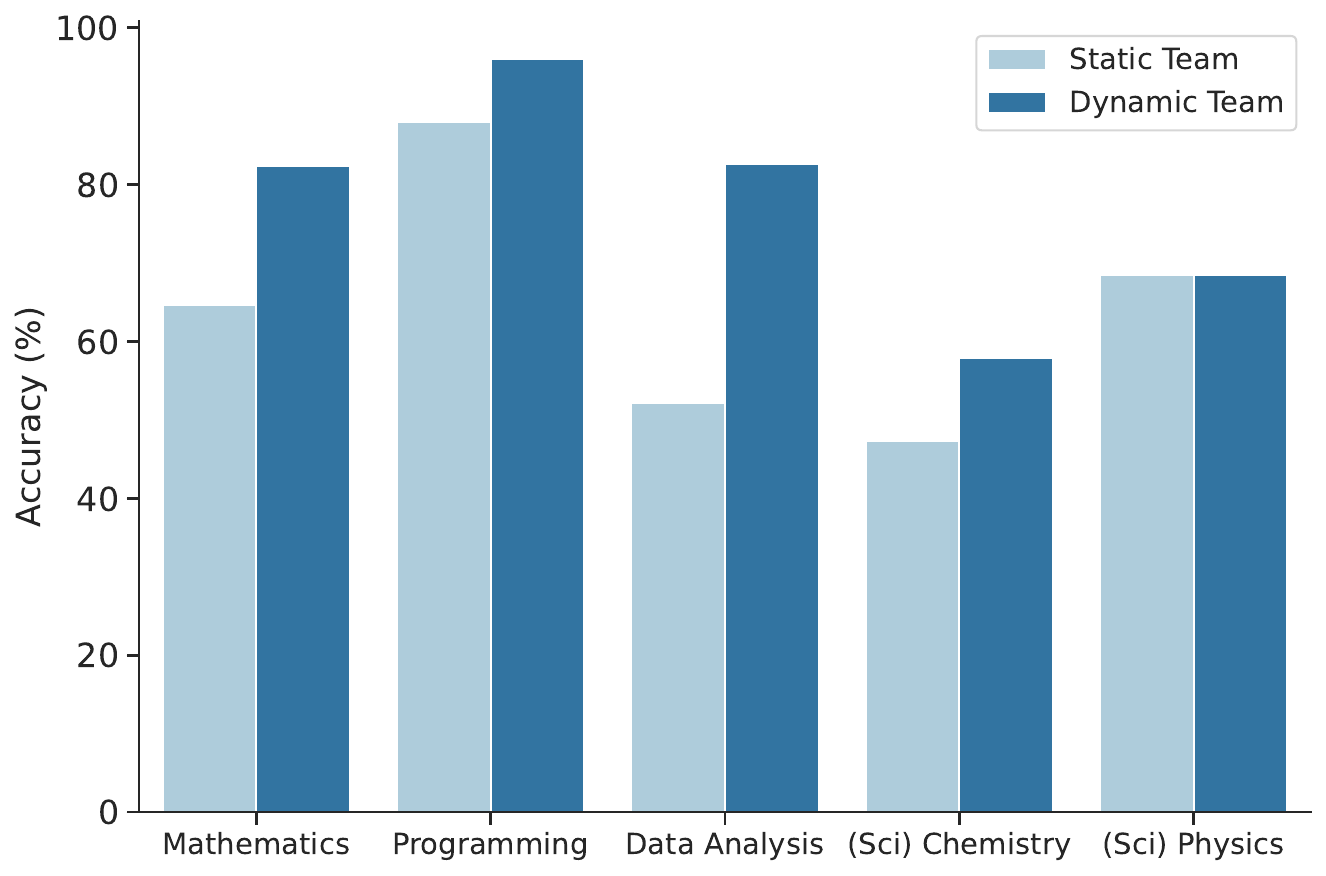}
    \caption{\label{fig:abla-team}Ablation comparison between static and adaptive team on the selected subset. Adaptive team during the conversation improves performance in different scenarios.}
\end{figure}

To further explore how building a team adaptively in conversation impacts the overall performance, we compare \Ours in two settings: (1) building a team at round 1 in the main conversation without changing the team members (\textit{static team}) and (2) adaptively changing team members and instructions across all rounds in the main conversation (\textit{adaptive team}). The comparison results in Figure~\ref{fig:abla-team} show that except in (Sci) Physics scenario, adaptive build outperforms static build across other scenarios. On data analysis scenarios, adaptive build outperforms static build by over 30.43\% on accuracy, showing the importance of changing team members when dealing with data analysis scenarios. 


\begin{table}[t]
\centering
\caption{\label{tab:ablation-gaia} Ablation study of tool library and agent library on information retrieval scenario (GAIA). We report the accuracy at each level and the average accuracy over three levels and mark the best results in \textbf{bold}.}
\resizebox{\columnwidth}{!}{%
\renewcommand\arraystretch{1.3}
\begin{tabular}{cccccc}
\hline \rowcolor{light-gray0} \multicolumn{2}{c}{\textbf{\Ours}} & \multicolumn{4}{c}{\textbf{Information Retrieval}} \\ \hline \rowcolor{light-gray}
\textbf{Agent Library}       & \textbf{Tool Library}       & \textbf{Level 1}        & \textbf{Level 2}        & \textbf{Level 3} & \textbf{Avg.}      \\ \hline
-                   & -                  &    32.07     &      13.95    &   3.84        & 18.18   \\
\checkmark          & -                  &    37.73       &    30.23        &  7.69          &  29.09   \\
-                   & \checkmark         &    39.62        &    19.78       &   7.69         &  24.24    \\
\checkmark          & \checkmark         & \textbf{56.60} & \textbf{39.53} & \textbf{11.54} & \textbf{40.60}     \\ \hline
\end{tabular}%
}
\end{table}

\begin{figure}[t!]
    \centering
    \includegraphics[width=\linewidth]{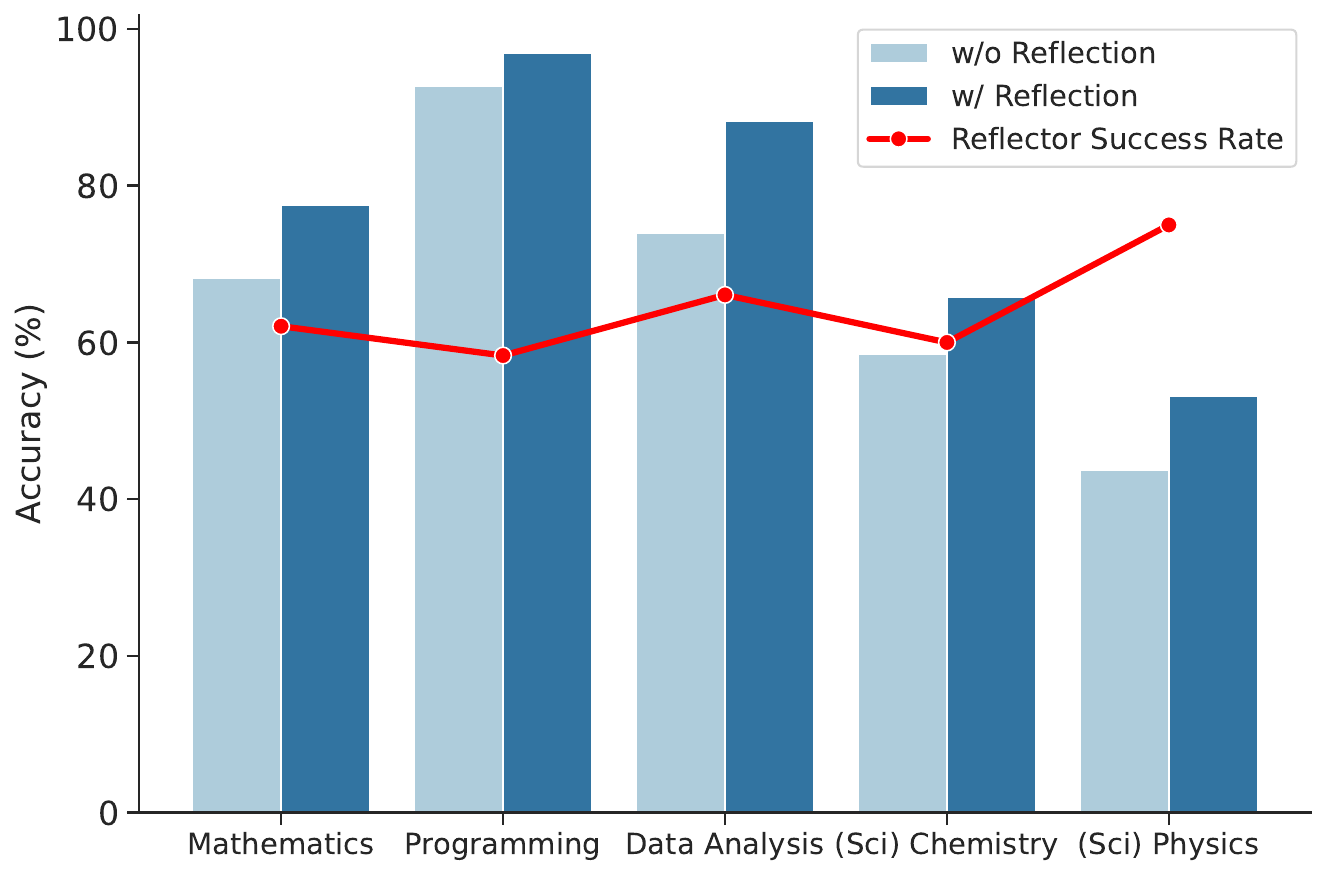}
    \vspace{-4mm}
    \caption{\label{fig:abla-ref}Ablation of reflection mechanism in \Ours. Reflector improves \Ours across all scenarios.}
\end{figure}

\subsubsection{Tool library and agent library} To demonstrate the effectiveness of agent and tool library, we conduct ablation studies over three settings: (1) remove both agent and tool library, (2) remove agent library while retaining the tool library, (3) remove tool library while retaining agent library, and (4) keep both agent and tool libraries. 
We selected GAIA as our target dataset because it represents the strong requirement for high-quality tools and the diversity of problem-solving abilities in different domains, which perfectly match the purpose of this ablation study.
As shown in Table \ref{tab:ablation-gaia}, removing the agent library and tool library can both significantly impair the system's performance. 
While both the tool and agent libraries can enhance performance independently, optimal results are achieved only when both libraries are employed concurrently. Handling level 1 tasks requires a moderate amount of web browsing and reasoning steps, which can be achieved by several single-turn tool calls or experts writing and executing code iteratively. 
Notably, without an agent library, \Ours performs much worse on Level 2 tasks. This is because these tasks are more sophisticated and mostly involve a significant number of web navigation and reasoning steps. Web browsing involves complex and dynamic interactions that are poorly suited to static tool libraries. The tasks require agents to coordinate multiple tools to solve them, which is a process prone to error in web scenarios filled with uncertainty. 

\subsubsection{Ablation on Reflector}
Reflection is one of the most important components in \Ours as its feedback represents the essence of the whole nested conversation and will, therefore, largely impact the decision-making process of \Ours. In Figure~\ref{fig:abla-ref}, we statistic the performance of \Ours with and without reflection and the success rate of the reflector. We define the \textit{success rate} as: if the LLM-based reflector can detect the dispute in nested conversation and highlight it for \Ours to avoid potential next-step error, we will mark it as "success." We calculate the success rate of the reflector across all problems under the corresponding scenario. We observe that in (sci) physics problems, the reflector represented a vital role with a high success rate and helped improve the performance of \Ours, but the performance of \Ours under physics remains lowest across all other scenarios. This is because the reflector cannot refine the problem-solving process but can detect the conflict inside a nested conversation and infer the possible error that may be incurred by the conflict. Therefore, the reflection mechanism helps prevent \textit{error propagation} in the multi-round problem-solving process designed for \Ours.

\subsubsection{Ablation on LLM Backbone and Cost Analysis}

\begin{figure}[t!]
    \centering
    \includegraphics[width=\linewidth]{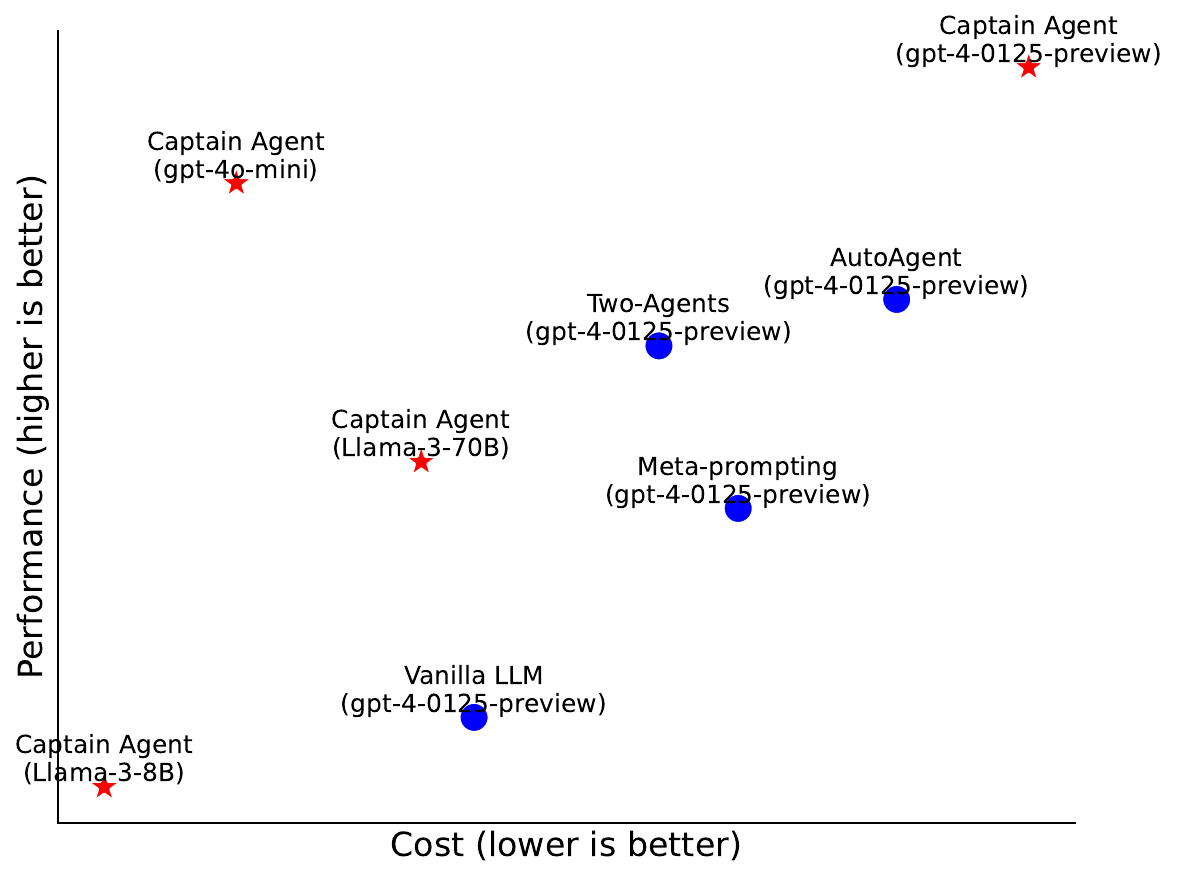}
    \vspace{-4mm}
    \caption{\label{fig:cost_perf}(Numerical results can be found in Table~\ref{tab:cost}) Comparison of performance on our reduced dataset for ablation study. \Ours achieves the best performance with \texttt{gpt-4-0125-preview}. \Ours with \texttt{gpt-4o-mini} can achieve competitive performance with other baselines that use \texttt{gpt-4-0125-preview}, and have significantly lower cost.}
\end{figure}

The high token cost associated with LLMs has always been a significant barrier to the practical deployment of agents, rendering them economically unfeasible. However, with the development of model distillation, the small model can also achieve competitive results with an agent framework on complex tasks.
To evaluate \Ours's ability to generalize to different LLM backbones for cost reduction, we equip \Ours and its nested experts with four different backbones, including \texttt{gpt-4-0125-preview}, \texttt{gpt-4o-mini}, \texttt{LLaMA-3-70B-Instruct}, and \texttt{LLaMA-3-8B-Instruct}, and compare it with all the baselines equipped with \texttt{gpt-4-0125-preview}.
Our cost statistic covers the whole \Ours workflow, including \Ours output, performing agent and tool selection, agent generation, and nested group conversation. 
We summarize the results based on the ranking of each setting in Figure~\ref{fig:cost_perf}. We observe that (1) \Ours with the same backbone as baselines can achieve the best performance but also the highest cost; (2) \Ours with \texttt{gpt-4o-mini} outperforms all other baselines but with a significantly lower cost, revealing the possibility of using a small language model as the backbone of an agent framework without seeing cost as a major issue.

\subsubsection{Agent Selected in Each Scenario}
\begin{figure*}[!t]
\centering
\subfigure[Mathematics]{
    \includegraphics[width=0.31\linewidth]{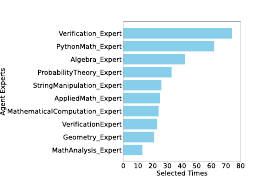}
}
\subfigure[Data Analysis]{
    \includegraphics[width=0.29\linewidth]{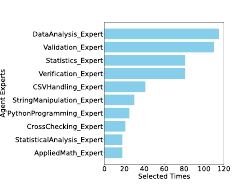}
}
\subfigure[Programming]{
    \includegraphics[width=0.29\linewidth]{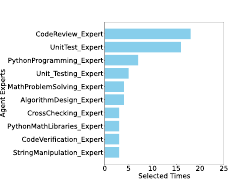}
}
\vspace{-2mm}
\subfigure[(Sci) Chemistry]{
    \includegraphics[width=0.31\linewidth]{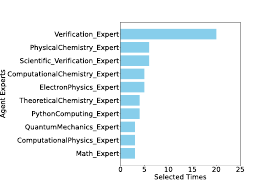}
}
\subfigure[(Sci) Physics]{
    \includegraphics[width=0.31\linewidth]{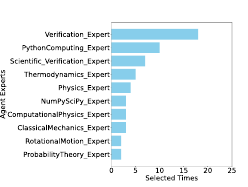}
}
\subfigure[Information retrieval]{
    \includegraphics[width=0.30\linewidth]{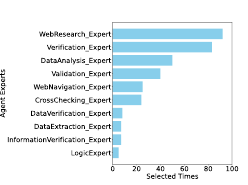}
}
\vspace{-2mm}
\caption{\label{fig:agent_freq} Top-10 selected agents and the corresponding selected times. We can observe that the selected agent is highly related to the scenario, and Verification\_Expert has a high selection rate.}
\end{figure*}

Our agent library records 541 agents in total after we complete all experiments, and we summarized the top-10 selected agent's frequency for each scenario in Figure~\ref{fig:agent_freq}. It is obvious that selected agents are highly related to the corresponding scenarios because we prompt \Ours in the system message to create a verification role to keep the conversation going smoothly. We also notice that in some scenarios (mathematics, data analysis, and programming), agents with a non-specific name and description will have a high selection rate (e.g., PythonMath\_Expert, DataAnalysis\_Expert, CodeReview\_Expert, etc.), as the embedding similarity between their description and user task is generally high. But we can still observe some specific agents being selected, like AppliedMath\_Expert, StringManipulation\_Expert, and NumPySciPy\_Expert, etc.
For the information retrieval scenario, WebResearch\_Expert has an extremely high selection rate. This agent is one of our hand-crafted agents who can search the internet and summarize a page in Markdown format, and Agent Filter can select this agent with \Ours instruction properly.

\section{Related Work}
Large language models (LLMs) represent a significant advancement in artificial intelligence, showcasing remarkable capabilities in various aspects, including reasoning~\citep{wei2022chain,yao2024tree,morishita2023learning, zhang2023ideal,li2023symbolic, ho2022large}, planning~\citep{babyagi,song2023llm, valmeekam2023planning, liu2023dynamic}, and adaptability to novel real-world observations~\citep{shi2024ehragent,hong2023metagpt,yang2023auto,dan2023educhat, zhou2023webarena,bharadhwaj2023roboagent}. Leveraging the inherent versatility of LLMs as generalized models adaptable to diverse scenarios, numerous efforts have been dedicated to the development of intelligent agents~\citep{wu2023autogen, xi2023rise, zhang2024training,sumers2023cognitive,zhou2023agents} where LLMs serve as foundational components.
For instance, one typical algorithm, React~\citep{yao2022react}, employs one single LLM to iteratively generate both reasoning trajectories and task-specific actions. This interleaved process enables the agent to engage in dynamic reasoning.
In addition, LLM agents can also harness external tools~\citep{qin2023tool,qin2023toolllm,schick2024toolformer,cai2023large,yuan2023craft,paranjape2023art,zhang2024training, huang2023metatool,ma2024m}, leveraging both their internal capabilities and external resources, collaborating effectively to solve more intricate problems.

The success of a single-agent system motivates the development of multiple-agent systems~\citep{wang2023survey, xi2023rise,autoagents, wu2023autogen, meta-prompting, hong2023metagpt, zhang2024training, zhang2023ecoassistant, valmeekam2023planning, wang2024tdag,saha2023branch,liang2023encouraging,du2023improving}. 
Methods focusing on static build require a protocol for agents to communicate with each other in a group chat and a builder that can receive the user's instruction and output an agent list~\citep{wu2023autogen,autoagents,hong2023metagpt}. The builder can be a human~\citep{wu2023autogen, hong2023metagpt} or a LLM agent~\citep{autoagents}. 
There are other works breaking down complex tasks into smaller components, each of which is then handled by a single specialized agent with detailed natural-language instructions~\citep{suzgun2024meta, zhuge2023mindstorms}. 
This task decomposition reduces the prediction burden on each agent by avoiding irrelevant context. For instance, meta-prompting~\citep{suzgun2024meta} involves a meta-model decomposing tasks and assigning subtasks to different LLMs for completion and aggregation. 
However, these methods lack a backtracking mechanism, 
making the problem-solving prototype fixed once agents are generated.
\vspace{-3mm}
\section{Conclusion and Discussion}
We introduce a new paradigm for multi-agent team-building, adaptive build. This new paradigm helps ensure diversity and prevent limited knowledge extraction. The new paradigm executed by our proposed agent, \Ours, manages agent teams for problem-solving steps using adaptive multi-agent team building and nested group conversation and reflection. Experimental results across six real-world scenarios demonstrate \Ours's efficacy in various tasks without prompt engineering, achieving superior results compared to existing methods. 
Ablation studies confirm that each component contributes equally to overall performance, underscoring the robustness of our approach.


\section*{Impact Statement}\label{sec:social-impact}
Our method dynamically ensembles LLM agents and equips them with versatile tools, allowing them to efficiently and effectively solve complex tasks. However, the development of agent systems that interact with the web environment raises safety concerns. The scope of our experiment in real-world interaction is limited to solving GAIA tasks, where the agents are required to search the web and browse websites. The agents are restricted from accessing publicly available information and are not capable of publishing content on the web. This ensures that our experiments remain both non-invasive and safe.

\bibliography{reference}

\begin{thebibliography}{81}
\providecommand{\natexlab}[1]{#1}
\providecommand{\url}[1]{\texttt{#1}}
\expandafter\ifx\csname urlstyle\endcsname\relax
  \providecommand{\doi}[1]{doi: #1}\else
  \providecommand{\doi}{doi: \begingroup \urlstyle{rm}\Url}\fi

\bibitem[AutoGenBench(2024)]{autogenbench}
AutoGenBench.
\newblock Github | autogenbench.
\newblock \url{https://microsoft.github.io/autogen/blog/2024/01/25/AutoGenBench}, 2024.

\bibitem[BabyAGI(2023)]{babyagi}
BabyAGI.
\newblock Github | babyagi.
\newblock \url{https://github.com/yoheinakajima/babyagi}, 2023.

\bibitem[Bharadhwaj et~al.(2023)Bharadhwaj, Vakil, Sharma, Gupta, Tulsiani, and Kumar]{bharadhwaj2023roboagent}
Bharadhwaj, H., Vakil, J., Sharma, M., Gupta, A., Tulsiani, S., and Kumar, V.
\newblock Roboagent: Generalization and efficiency in robot manipulation via semantic augmentations and action chunking.
\newblock \emph{arXiv preprint arXiv:2309.01918}, 2023.

\bibitem[Brown et~al.(2020)Brown, Mann, Ryder, Subbiah, Kaplan, Dhariwal, Neelakantan, Shyam, Sastry, Askell, et~al.]{brown2020language}
Brown, T., Mann, B., Ryder, N., Subbiah, M., Kaplan, J.~D., Dhariwal, P., Neelakantan, A., Shyam, P., Sastry, G., Askell, A., et~al.
\newblock Language models are few-shot learners.
\newblock \emph{Advances in neural information processing systems}, 33:\penalty0 1877--1901, 2020.

\bibitem[Cai et~al.(2023)Cai, Wang, Ma, Chen, and Zhou]{cai2023large}
Cai, T., Wang, X., Ma, T., Chen, X., and Zhou, D.
\newblock Large language models as tool makers.
\newblock \emph{arXiv preprint arXiv:2305.17126}, 2023.

\bibitem[Chen et~al.(2023)Chen, Dong, Shu, Zhang, Sesay, Karlsson, Fu, and Shi]{autoagents}
Chen, G., Dong, S., Shu, Y., Zhang, G., Sesay, J., Karlsson, B.~F., Fu, J., and Shi, Y.
\newblock Autoagents: A framework for automatic agent generation.
\newblock \emph{arXiv preprint arXiv:2309.17288}, 2023.

\bibitem[Chen et~al.(2021)Chen, Tworek, Jun, Yuan, de~Oliveira~Pinto, Kaplan, Edwards, Burda, Joseph, Brockman, Ray, Puri, Krueger, Petrov, Khlaaf, Sastry, Mishkin, Chan, Gray, Ryder, Pavlov, Power, Kaiser, Bavarian, Winter, Tillet, Such, Cummings, Plappert, Chantzis, Barnes, Herbert-Voss, Guss, Nichol, Paino, Tezak, Tang, Babuschkin, Balaji, Jain, Saunders, Hesse, Carr, Leike, Achiam, Misra, Morikawa, Radford, Knight, Brundage, Murati, Mayer, Welinder, McGrew, Amodei, McCandlish, Sutskever, and Zaremba]{chen2021codex}
Chen, M., Tworek, J., Jun, H., Yuan, Q., de~Oliveira~Pinto, H.~P., Kaplan, J., Edwards, H., Burda, Y., Joseph, N., Brockman, G., Ray, A., Puri, R., Krueger, G., Petrov, M., Khlaaf, H., Sastry, G., Mishkin, P., Chan, B., Gray, S., Ryder, N., Pavlov, M., Power, A., Kaiser, L., Bavarian, M., Winter, C., Tillet, P., Such, F.~P., Cummings, D., Plappert, M., Chantzis, F., Barnes, E., Herbert-Voss, A., Guss, W.~H., Nichol, A., Paino, A., Tezak, N., Tang, J., Babuschkin, I., Balaji, S., Jain, S., Saunders, W., Hesse, C., Carr, A.~N., Leike, J., Achiam, J., Misra, V., Morikawa, E., Radford, A., Knight, M., Brundage, M., Murati, M., Mayer, K., Welinder, P., McGrew, B., Amodei, D., McCandlish, S., Sutskever, I., and Zaremba, W.
\newblock Evaluating large language models trained on code.
\newblock 2021.

\bibitem[Chen et~al.(2024)Chen, Su, Zuo, Yang, Yuan, Chan, Yu, Lu, Hung, Qian, Qin, Cong, Xie, Liu, Sun, and Zhou]{chen2024agentverse}
Chen, W., Su, Y., Zuo, J., Yang, C., Yuan, C., Chan, C.-M., Yu, H., Lu, Y., Hung, Y.-H., Qian, C., Qin, Y., Cong, X., Xie, R., Liu, Z., Sun, M., and Zhou, J.
\newblock Agentverse: Facilitating multi-agent collaboration and exploring emergent behaviors.
\newblock In \emph{The Twelfth International Conference on Learning Representations}, 2024.
\newblock URL \url{https://openreview.net/forum?id=EHg5GDnyq1}.

\bibitem[Confer et~al.(2010)Confer, Easton, Fleischman, Goetz, Lewis, Perilloux, and Buss]{confer2010evolutionary}
Confer, J.~C., Easton, J.~A., Fleischman, D.~S., Goetz, C.~D., Lewis, D.~M., Perilloux, C., and Buss, D.~M.
\newblock Evolutionary psychology: Controversies, questions, prospects, and limitations.
\newblock \emph{American psychologist}, 65\penalty0 (2):\penalty0 110, 2010.

\bibitem[Dai et~al.(2023)Dai, Sun, Dong, Hao, Ma, Sui, and Wei]{dai2023can}
Dai, D., Sun, Y., Dong, L., Hao, Y., Ma, S., Sui, Z., and Wei, F.
\newblock Why can gpt learn in-context? language models secretly perform gradient descent as meta-optimizers.
\newblock In \emph{Findings of the Association for Computational Linguistics: ACL 2023}, pp.\  4005--4019, 2023.

\bibitem[Dan et~al.(2023)Dan, Lei, Gu, Li, Yin, Lin, Ye, Tie, Zhou, Wang, et~al.]{dan2023educhat}
Dan, Y., Lei, Z., Gu, Y., Li, Y., Yin, J., Lin, J., Ye, L., Tie, Z., Zhou, Y., Wang, Y., et~al.
\newblock Educhat: A large-scale language model-based chatbot system for intelligent education.
\newblock \emph{arXiv preprint arXiv:2308.02773}, 2023.

\bibitem[Dong et~al.(2022)Dong, Li, Dai, Zheng, Wu, Chang, Sun, Xu, and Sui]{dong2022survey}
Dong, Q., Li, L., Dai, D., Zheng, C., Wu, Z., Chang, B., Sun, X., Xu, J., and Sui, Z.
\newblock A survey on in-context learning.
\newblock \emph{arXiv preprint arXiv:2301.00234}, 2022.

\bibitem[Du et~al.(2023)Du, Li, Torralba, Tenenbaum, and Mordatch]{du2023improving}
Du, Y., Li, S., Torralba, A., Tenenbaum, J.~B., and Mordatch, I.
\newblock Improving factuality and reasoning in language models through multiagent debate.
\newblock \emph{arXiv preprint arXiv:2305.14325}, 2023.

\bibitem[Elimari \& Lafargue(2020)Elimari and Lafargue]{elimari2020network}
Elimari, N. and Lafargue, G.
\newblock Network neuroscience and the adapted mind: Rethinking the role of network theories in evolutionary psychology.
\newblock \emph{Frontiers in psychology}, 11:\penalty0 545632, 2020.

\bibitem[Fernandes et~al.(2023)Fernandes, Madaan, Liu, Farinhas, Martins, Bertsch, de~Souza, Zhou, Wu, Neubig, et~al.]{fernandes2023bridging}
Fernandes, P., Madaan, A., Liu, E., Farinhas, A., Martins, P.~H., Bertsch, A., de~Souza, J.~G., Zhou, S., Wu, T., Neubig, G., et~al.
\newblock Bridging the gap: A survey on integrating (human) feedback for natural language generation.
\newblock \emph{Transactions of the Association for Computational Linguistics}, 11:\penalty0 1643--1668, 2023.

\bibitem[Furuta et~al.(2024)Furuta, Lee, Nachum, Matsuo, Faust, Gu, and Gur]{furuta2024multimodal}
Furuta, H., Lee, K.-H., Nachum, O., Matsuo, Y., Faust, A., Gu, S.~S., and Gur, I.
\newblock Multimodal web navigation with instruction-finetuned foundation models.
\newblock In \emph{The Twelfth International Conference on Learning Representations}, 2024.
\newblock URL \url{https://openreview.net/forum?id=efFmBWioSc}.

\bibitem[GAIA\_Orchestrator(2024)]{orchestrator}
GAIA\_Orchestrator.
\newblock Github | autogen: Gaia orchestrator.
\newblock \url{https://github.com/microsoft/autogen/tree/gaia_multiagent_v01_march_1st/samples/tools/autogenbench/scenarios/GAIA/Templates/Orchestrator}, 2024.

\bibitem[Gao et~al.(2023)Gao, Xiong, Gao, Jia, Pan, Bi, Dai, Sun, and Wang]{gao2023retrieval}
Gao, Y., Xiong, Y., Gao, X., Jia, K., Pan, J., Bi, Y., Dai, Y., Sun, J., and Wang, H.
\newblock Retrieval-augmented generation for large language models: A survey.
\newblock \emph{arXiv preprint arXiv:2312.10997}, 2023.

\bibitem[Hendrycks et~al.(2021{\natexlab{a}})Hendrycks, Burns, Kadavath, Arora, Basart, Tang, Song, and Steinhardt]{hendrycksmath2021}
Hendrycks, D., Burns, C., Kadavath, S., Arora, A., Basart, S., Tang, E., Song, D., and Steinhardt, J.
\newblock Measuring mathematical problem solving with the math dataset.
\newblock \emph{NeurIPS}, 2021{\natexlab{a}}.

\bibitem[Hendrycks et~al.(2021{\natexlab{b}})Hendrycks, Burns, Kadavath, Arora, Basart, Tang, Song, and Steinhardt]{math}
Hendrycks, D., Burns, C., Kadavath, S., Arora, A., Basart, S., Tang, E., Song, D., and Steinhardt, J.
\newblock Measuring mathematical problem solving with the math dataset.
\newblock In \emph{Thirty-fifth Conference on Neural Information Processing Systems Datasets and Benchmarks Track (Round 2)}, 2021{\natexlab{b}}.

\bibitem[Ho et~al.(2022)Ho, Schmid, and Yun]{ho2022large}
Ho, N., Schmid, L., and Yun, S.-Y.
\newblock Large language models are reasoning teachers.
\newblock \emph{arXiv preprint arXiv:2212.10071}, 2022.

\bibitem[Hong et~al.(2023)Hong, Zheng, Chen, Cheng, Wang, Zhang, Wang, Yau, Lin, Zhou, et~al.]{hong2023metagpt}
Hong, S., Zheng, X., Chen, J., Cheng, Y., Wang, J., Zhang, C., Wang, Z., Yau, S. K.~S., Lin, Z., Zhou, L., et~al.
\newblock Metagpt: Meta programming for multi-agent collaborative framework.
\newblock \emph{arXiv preprint arXiv:2308.00352}, 2023.

\bibitem[Hong et~al.(2024)Hong, Lin, Liu, Wu, Li, Chen, Zhang, Wang, Zhang, Zhuge, et~al.]{datainterpreter}
Hong, S., Lin, Y., Liu, B., Wu, B., Li, D., Chen, J., Zhang, J., Wang, J., Zhang, L., Zhuge, M., et~al.
\newblock Data interpreter: An llm agent for data science.
\newblock \emph{arXiv preprint arXiv:2402.18679}, 2024.

\bibitem[Hu et~al.(2024{\natexlab{a}})Hu, Zhao, Wei, Chai, Ma, Wang, Wang, Su, Xu, Zhu, Cheng, Yuan, Li, Kuang, Yang, Yang, and Wu]{hu2024infiagentdabench}
Hu, X., Zhao, Z., Wei, S., Chai, Z., Ma, Q., Wang, G., Wang, X., Su, J., Xu, J., Zhu, M., Cheng, Y., Yuan, J., Li, J., Kuang, K., Yang, Y., Yang, H., and Wu, F.
\newblock Infiagent-dabench: Evaluating agents on data analysis tasks, 2024{\natexlab{a}}.

\bibitem[Hu et~al.(2024{\natexlab{b}})Hu, Zhao, Wei, Chai, Wang, Wang, Su, Xu, Zhu, Cheng, et~al.]{dabench}
Hu, X., Zhao, Z., Wei, S., Chai, Z., Wang, G., Wang, X., Su, J., Xu, J., Zhu, M., Cheng, Y., et~al.
\newblock Infiagent-dabench: Evaluating agents on data analysis tasks.
\newblock \emph{arXiv preprint arXiv:2401.05507}, 2024{\natexlab{b}}.

\bibitem[Huang et~al.(2023)Huang, Shi, Li, Fan, Wu, Zhang, Liu, Zhou, Wan, Gong, et~al.]{huang2023metatool}
Huang, Y., Shi, J., Li, Y., Fan, C., Wu, S., Zhang, Q., Liu, Y., Zhou, P., Wan, Y., Gong, N.~Z., et~al.
\newblock Metatool benchmark for large language models: Deciding whether to use tools and which to use.
\newblock \emph{arXiv preprint arXiv:2310.03128}, 2023.

\bibitem[Le et~al.(2020)Le, Chen, and Babar]{humaneval}
Le, T.~H., Chen, H., and Babar, M.~A.
\newblock Deep learning for source code modeling and generation: Models, applications, and challenges.
\newblock \emph{ACM Computing Surveys (CSUR)}, 53\penalty0 (3):\penalty0 1--38, 2020.

\bibitem[Lewis et~al.(2020)Lewis, Perez, Piktus, Petroni, Karpukhin, Goyal, K{\"u}ttler, Lewis, Yih, Rockt{\"a}schel, et~al.]{lewis2020retrieval}
Lewis, P., Perez, E., Piktus, A., Petroni, F., Karpukhin, V., Goyal, N., K{\"u}ttler, H., Lewis, M., Yih, W.-t., Rockt{\"a}schel, T., et~al.
\newblock Retrieval-augmented generation for knowledge-intensive nlp tasks.
\newblock \emph{Advances in Neural Information Processing Systems}, 33:\penalty0 9459--9474, 2020.

\bibitem[Li et~al.(2023{\natexlab{a}})Li, Hessel, Yu, Ren, Chang, and Choi]{li2023symbolic}
Li, L.~H., Hessel, J., Yu, Y., Ren, X., Chang, K.-W., and Choi, Y.
\newblock Symbolic chain-of-thought distillation: Small models can also" think" step-by-step.
\newblock \emph{arXiv preprint arXiv:2306.14050}, 2023{\natexlab{a}}.

\bibitem[Li et~al.(2023{\natexlab{b}})Li, Gao, Li, and Liao]{li2023large}
Li, N., Gao, C., Li, Y., and Liao, Q.
\newblock Large language model-empowered agents for simulating macroeconomic activities.
\newblock \emph{Available at SSRN 4606937}, 2023{\natexlab{b}}.

\bibitem[Li et~al.(2023{\natexlab{c}})Li, Ildiz, Papailiopoulos, and Oymak]{li2023transformers}
Li, Y., Ildiz, M.~E., Papailiopoulos, D., and Oymak, S.
\newblock Transformers as algorithms: Generalization and stability in in-context learning.
\newblock In \emph{International Conference on Machine Learning}, pp.\  19565--19594. PMLR, 2023{\natexlab{c}}.

\bibitem[Liang et~al.(2023)Liang, He, Jiao, Wang, Wang, Wang, Yang, Tu, and Shi]{liang2023encouraging}
Liang, T., He, Z., Jiao, W., Wang, X., Wang, Y., Wang, R., Yang, Y., Tu, Z., and Shi, S.
\newblock Encouraging divergent thinking in large language models through multi-agent debate.
\newblock \emph{arXiv preprint arXiv:2305.19118}, 2023.

\bibitem[Liu et~al.(2023{\natexlab{a}})Liu, Jiang, Zhang, Liu, Zhang, Biswas, and Stone]{Liu2023LLMPEL}
Liu, B., Jiang, Y., Zhang, X., Liu, Q., Zhang, S., Biswas, J., and Stone, P.
\newblock Llm+p: Empowering large language models with optimal planning proficiency.
\newblock \emph{ArXiv}, abs/2304.11477, 2023{\natexlab{a}}.
\newblock URL \url{https://api.semanticscholar.org/CorpusID:258298051}.

\bibitem[Liu et~al.(2023{\natexlab{b}})Liu, Zhang, Li, Liu, and Yang]{liu2023dynamic}
Liu, Z., Zhang, Y., Li, P., Liu, Y., and Yang, D.
\newblock Dynamic llm-agent network: An llm-agent collaboration framework with agent team optimization.
\newblock \emph{arXiv preprint arXiv:2310.02170}, 2023{\natexlab{b}}.

\bibitem[Ma et~al.(2024)Ma, Huang, Zhang, Gupta, and Krishna]{ma2024m}
Ma, Z., Huang, W., Zhang, J., Gupta, T., and Krishna, R.
\newblock m\&m's: A benchmark to evaluate tool-use for multi-step multi-modal tasks.
\newblock In \emph{Synthetic Data for Computer Vision Workshop@ CVPR 2024}, 2024.

\bibitem[Mao et~al.(2016)Mao, Mason, Suri, and Watts]{mao2016experimental}
Mao, A., Mason, W., Suri, S., and Watts, D.~J.
\newblock An experimental study of team size and performance on a complex task.
\newblock \emph{PloS one}, 11\penalty0 (4):\penalty0 e0153048, 2016.

\bibitem[Mialon et~al.(2023)Mialon, Fourrier, Swift, Wolf, LeCun, and Scialom]{mialon2023gaia}
Mialon, G., Fourrier, C., Swift, C., Wolf, T., LeCun, Y., and Scialom, T.
\newblock Gaia: a benchmark for general ai assistants.
\newblock \emph{arXiv preprint arXiv:2311.12983}, 2023.

\bibitem[Mialon et~al.(2024)Mialon, Fourrier, Wolf, LeCun, and Scialom]{gaia}
Mialon, G., Fourrier, C., Wolf, T., LeCun, Y., and Scialom, T.
\newblock {GAIA}: a benchmark for general {AI} assistants.
\newblock In \emph{The Twelfth International Conference on Learning Representations}, 2024.
\newblock URL \url{https://openreview.net/forum?id=fibxvahvs3}.

\bibitem[Morishita et~al.(2023)Morishita, Morio, Yamaguchi, and Sogawa]{morishita2023learning}
Morishita, T., Morio, G., Yamaguchi, A., and Sogawa, Y.
\newblock Learning deductive reasoning from synthetic corpus based on formal logic.
\newblock In \emph{International Conference on Machine Learning}, pp.\  25254--25274. PMLR, 2023.

\bibitem[Paranjape et~al.(2023)Paranjape, Lundberg, Singh, Hajishirzi, Zettlemoyer, and Ribeiro]{paranjape2023art}
Paranjape, B., Lundberg, S., Singh, S., Hajishirzi, H., Zettlemoyer, L., and Ribeiro, M.~T.
\newblock Art: Automatic multi-step reasoning and tool-use for large language models.
\newblock \emph{arXiv preprint arXiv:2303.09014}, 2023.

\bibitem[Qin et~al.(2023{\natexlab{a}})Qin, Hu, Lin, Chen, Ding, Cui, Zeng, Huang, Xiao, Han, Fung, Su, Wang, Qian, Tian, Zhu, Liang, Shen, Xu, Zhang, Ye, Li, Tang, Yi, Zhu, Dai, Yan, Cong, Lu, Zhao, Huang, Yan, Han, Sun, Li, Phang, Yang, Wu, Ji, Liu, and Sun]{qin2023tool}
Qin, Y., Hu, S., Lin, Y., Chen, W., Ding, N., Cui, G., Zeng, Z., Huang, Y., Xiao, C., Han, C., Fung, Y.~R., Su, Y., Wang, H., Qian, C., Tian, R., Zhu, K., Liang, S., Shen, X., Xu, B., Zhang, Z., Ye, Y., Li, B., Tang, Z., Yi, J., Zhu, Y., Dai, Z., Yan, L., Cong, X., Lu, Y., Zhao, W., Huang, Y., Yan, J., Han, X., Sun, X., Li, D., Phang, J., Yang, C., Wu, T., Ji, H., Liu, Z., and Sun, M.
\newblock Tool learning with foundation models, 2023{\natexlab{a}}.

\bibitem[Qin et~al.(2023{\natexlab{b}})Qin, Liang, Ye, Zhu, Yan, Lu, Lin, Cong, Tang, Qian, Zhao, Hong, Tian, Xie, Zhou, Gerstein, Li, Liu, and Sun]{qin2023toolllm}
Qin, Y., Liang, S., Ye, Y., Zhu, K., Yan, L., Lu, Y., Lin, Y., Cong, X., Tang, X., Qian, B., Zhao, S., Hong, L., Tian, R., Xie, R., Zhou, J., Gerstein, M., Li, D., Liu, Z., and Sun, M.
\newblock Toolllm: Facilitating large language models to master 16000+ real-world apis, 2023{\natexlab{b}}.

\bibitem[Ram et~al.(2023)Ram, Levine, Dalmedigos, Muhlgay, Shashua, Leyton-Brown, and Shoham]{ram2023context}
Ram, O., Levine, Y., Dalmedigos, I., Muhlgay, D., Shashua, A., Leyton-Brown, K., and Shoham, Y.
\newblock In-context retrieval-augmented language models.
\newblock \emph{Transactions of the Association for Computational Linguistics}, 11:\penalty0 1316--1331, 2023.

\bibitem[Saha et~al.(2023)Saha, Levy, Celikyilmaz, Bansal, Weston, and Li]{saha2023branch}
Saha, S., Levy, O., Celikyilmaz, A., Bansal, M., Weston, J., and Li, X.
\newblock Branch-solve-merge improves large language model evaluation and generation.
\newblock \emph{arXiv preprint arXiv:2310.15123}, 2023.

\bibitem[Schick et~al.(2024)Schick, Dwivedi-Yu, Dess{\`\i}, Raileanu, Lomeli, Hambro, Zettlemoyer, Cancedda, and Scialom]{schick2024toolformer}
Schick, T., Dwivedi-Yu, J., Dess{\`\i}, R., Raileanu, R., Lomeli, M., Hambro, E., Zettlemoyer, L., Cancedda, N., and Scialom, T.
\newblock Toolformer: Language models can teach themselves to use tools.
\newblock \emph{Advances in Neural Information Processing Systems}, 36, 2024.

\bibitem[Shi et~al.(2024)Shi, Xu, Zhuang, Yu, Zhang, Wu, Zhu, Ho, Yang, and Wang]{shi2024ehragent}
Shi, W., Xu, R., Zhuang, Y., Yu, Y., Zhang, J., Wu, H., Zhu, Y., Ho, J., Yang, C., and Wang, M.~D.
\newblock Ehragent: Code empowers large language models for complex tabular reasoning on electronic health records.
\newblock \emph{arXiv preprint arXiv:2401.07128}, 2024.

\bibitem[Shinn et~al.(2024)Shinn, Cassano, Gopinath, Narasimhan, and Yao]{shinn2024reflexion}
Shinn, N., Cassano, F., Gopinath, A., Narasimhan, K., and Yao, S.
\newblock Reflexion: Language agents with verbal reinforcement learning.
\newblock \emph{Advances in Neural Information Processing Systems}, 36, 2024.

\bibitem[Song et~al.(2023)Song, Wu, Washington, Sadler, Chao, and Su]{song2023llm}
Song, C.~H., Wu, J., Washington, C., Sadler, B.~M., Chao, W.-L., and Su, Y.
\newblock Llm-planner: Few-shot grounded planning for embodied agents with large language models.
\newblock In \emph{Proceedings of the IEEE/CVF International Conference on Computer Vision}, pp.\  2998--3009, 2023.

\bibitem[Sumers et~al.(2023)Sumers, Yao, Narasimhan, and Griffiths]{sumers2023cognitive}
Sumers, T.~R., Yao, S., Narasimhan, K., and Griffiths, T.~L.
\newblock Cognitive architectures for language agents.
\newblock \emph{arXiv preprint arXiv:2309.02427}, 2023.

\bibitem[Sun et~al.(2024)Sun, Zhuang, Kong, Dai, and Zhang]{sun2024adaplanner}
Sun, H., Zhuang, Y., Kong, L., Dai, B., and Zhang, C.
\newblock Adaplanner: Adaptive planning from feedback with language models.
\newblock \emph{Advances in Neural Information Processing Systems}, 36, 2024.

\bibitem[Suzgun \& Kalai(2024{\natexlab{a}})Suzgun and Kalai]{meta-prompting}
Suzgun, M. and Kalai, A.~T.
\newblock Meta-prompting: Enhancing language models with task-agnostic scaffolding.
\newblock \emph{arXiv preprint arXiv:2401.12954}, 2024{\natexlab{a}}.

\bibitem[Suzgun \& Kalai(2024{\natexlab{b}})Suzgun and Kalai]{suzgun2024meta}
Suzgun, M. and Kalai, A.~T.
\newblock Meta-prompting: Enhancing language models with task-agnostic scaffolding.
\newblock \emph{arXiv preprint arXiv:2401.12954}, 2024{\natexlab{b}}.

\bibitem[Valmeekam et~al.(2022)Valmeekam, Olmo, Sreedharan, and Kambhampati]{Valmeekam2022PlanBenchAE}
Valmeekam, K., Olmo, A., Sreedharan, S., and Kambhampati, S.
\newblock Planbench: An extensible benchmark for evaluating large language models on planning and reasoning about change.
\newblock In \emph{Neural Information Processing Systems}, 2022.
\newblock URL \url{https://api.semanticscholar.org/CorpusID:249889477}.

\bibitem[Valmeekam et~al.(2023)Valmeekam, Marquez, Sreedharan, and Kambhampati]{valmeekam2023planning}
Valmeekam, K., Marquez, M., Sreedharan, S., and Kambhampati, S.
\newblock On the planning abilities of large language models-a critical investigation.
\newblock \emph{Advances in Neural Information Processing Systems}, 36:\penalty0 75993--76005, 2023.

\bibitem[Wang et~al.(2023{\natexlab{a}})Wang, Ma, Feng, Zhang, Yang, Zhang, Chen, Tang, Chen, Lin, et~al.]{wang2023survey}
Wang, L., Ma, C., Feng, X., Zhang, Z., Yang, H., Zhang, J., Chen, Z., Tang, J., Chen, X., Lin, Y., et~al.
\newblock A survey on large language model based autonomous agents.
\newblock \emph{arXiv preprint arXiv:2308.11432}, 2023{\natexlab{a}}.

\bibitem[Wang et~al.(2023{\natexlab{b}})Wang, Hu, Lu, Zhu, Zhang, Subramaniam, Loomba, Zhang, Sun, and Wang]{scibench}
Wang, X., Hu, Z., Lu, P., Zhu, Y., Zhang, J., Subramaniam, S., Loomba, A.~R., Zhang, S., Sun, Y., and Wang, W.
\newblock Scibench: Evaluating college-level scientific problem-solving abilities of large language models.
\newblock \emph{arXiv preprint arXiv:2307.10635}, 2023{\natexlab{b}}.

\bibitem[Wang et~al.(2023{\natexlab{c}})Wang, Wang, Liu, Chen, Yuan, Peng, and Ji]{wang2023mint}
Wang, X., Wang, Z., Liu, J., Chen, Y., Yuan, L., Peng, H., and Ji, H.
\newblock Mint: Evaluating llms in multi-turn interaction with tools and language feedback.
\newblock \emph{arXiv preprint arXiv:2309.10691}, 2023{\natexlab{c}}.

\bibitem[Wang et~al.(2024)Wang, Wu, Yao, and Su]{wang2024tdag}
Wang, Y., Wu, Z., Yao, J., and Su, J.
\newblock Tdag: A multi-agent framework based on dynamic task decomposition and agent generation.
\newblock \emph{arXiv preprint arXiv:2402.10178}, 2024.

\bibitem[Wei et~al.(2022{\natexlab{a}})Wei, Wang, Schuurmans, Bosma, hsin Chi, Xia, Le, and Zhou]{Wei2022ChainOT}
Wei, J., Wang, X., Schuurmans, D., Bosma, M., hsin Chi, E.~H., Xia, F., Le, Q., and Zhou, D.
\newblock Chain of thought prompting elicits reasoning in large language models.
\newblock \emph{ArXiv}, abs/2201.11903, 2022{\natexlab{a}}.
\newblock URL \url{https://api.semanticscholar.org/CorpusID:246411621}.

\bibitem[Wei et~al.(2022{\natexlab{b}})Wei, Wang, Schuurmans, Bosma, Xia, Chi, Le, Zhou, et~al.]{wei2022chain}
Wei, J., Wang, X., Schuurmans, D., Bosma, M., Xia, F., Chi, E., Le, Q.~V., Zhou, D., et~al.
\newblock Chain-of-thought prompting elicits reasoning in large language models.
\newblock \emph{Advances in neural information processing systems}, 35:\penalty0 24824--24837, 2022{\natexlab{b}}.

\bibitem[Wu et~al.(2023)Wu, Bansal, Zhang, Wu, Zhang, Zhu, Li, Jiang, Zhang, and Wang]{wu2023autogen}
Wu, Q., Bansal, G., Zhang, J., Wu, Y., Zhang, S., Zhu, E., Li, B., Jiang, L., Zhang, X., and Wang, C.
\newblock Autogen: Enabling next-gen llm applications via multi-agent conversation framework.
\newblock \emph{arXiv preprint arXiv:2308.08155}, 2023.

\bibitem[Wu et~al.(2024)Wu, Han, Ding, Weng, Liu, Yao, Yu, and Kong]{friday}
Wu, Z., Han, C., Ding, Z., Weng, Z., Liu, Z., Yao, S., Yu, T., and Kong, L.
\newblock Os-copilot: Towards generalist computer agents with self-improvement.
\newblock \emph{arXiv preprint arXiv:2402.07456}, 2024.

\bibitem[Xi et~al.(2023)Xi, Chen, Guo, He, Ding, Hong, Zhang, Wang, Jin, Zhou, et~al.]{xi2023rise}
Xi, Z., Chen, W., Guo, X., He, W., Ding, Y., Hong, B., Zhang, M., Wang, J., Jin, S., Zhou, E., et~al.
\newblock The rise and potential of large language model based agents: A survey.
\newblock \emph{arXiv preprint arXiv:2309.07864}, 2023.

\bibitem[Xie et~al.(2024)Xie, Zhang, Chen, Zhu, Lou, Tian, Xiao, and Su]{xie2024travelplanner}
Xie, J., Zhang, K., Chen, J., Zhu, T., Lou, R., Tian, Y., Xiao, Y., and Su, Y.
\newblock Travelplanner: A benchmark for real-world planning with language agents.
\newblock \emph{arXiv preprint arXiv:2402.01622}, 2024.

\bibitem[Xu et~al.(2024)Xu, Wang, Fan, and Liu]{xu2024benchmarking}
Xu, R., Wang, Z., Fan, R.-Z., and Liu, P.
\newblock Benchmarking benchmark leakage in large language models.
\newblock \emph{arXiv preprint arXiv:2404.18824}, 2024.

\bibitem[Yang et~al.(2023{\natexlab{a}})Yang, Yue, and He]{yang2023auto}
Yang, H., Yue, S., and He, Y.
\newblock Auto-gpt for online decision making: Benchmarks and additional opinions.
\newblock \emph{arXiv preprint arXiv:2306.02224}, 2023{\natexlab{a}}.

\bibitem[Yang et~al.(2023{\natexlab{b}})Yang, Hui, Yang, Li, Huang, and Li]{yang2023iterative}
Yang, J., Hui, B., Yang, M., Li, B., Huang, F., and Li, Y.
\newblock Iterative forward tuning boosts in-context learning in language models.
\newblock \emph{arXiv preprint arXiv:2305.13016}, 2023{\natexlab{b}}.

\bibitem[Yang et~al.(2024{\natexlab{a}})Yang, Jimenez, Wettig, Lieret, Yao, Narasimhan, and Press]{yang2024sweagent}
Yang, J., Jimenez, C.~E., Wettig, A., Lieret, K., Yao, S., Narasimhan, K., and Press, O.
\newblock Swe-agent: Agent computer interfaces enable software engineering language models, 2024{\natexlab{a}}.

\bibitem[Yang et~al.(2024{\natexlab{b}})Yang, Prabhakar, Narasimhan, and Yao]{yang2024intercode}
Yang, J., Prabhakar, A., Narasimhan, K., and Yao, S.
\newblock Intercode: Standardizing and benchmarking interactive coding with execution feedback.
\newblock \emph{Advances in Neural Information Processing Systems}, 36, 2024{\natexlab{b}}.

\bibitem[Yao et~al.(2022)Yao, Zhao, Yu, Du, Shafran, Narasimhan, and Cao]{yao2022react}
Yao, S., Zhao, J., Yu, D., Du, N., Shafran, I., Narasimhan, K., and Cao, Y.
\newblock React: Synergizing reasoning and acting in language models.
\newblock \emph{arXiv preprint arXiv:2210.03629}, 2022.

\bibitem[Yao et~al.(2024)Yao, Yu, Zhao, Shafran, Griffiths, Cao, and Narasimhan]{yao2024tree}
Yao, S., Yu, D., Zhao, J., Shafran, I., Griffiths, T., Cao, Y., and Narasimhan, K.
\newblock Tree of thoughts: Deliberate problem solving with large language models.
\newblock \emph{Advances in Neural Information Processing Systems}, 36, 2024.

\bibitem[Yuan et~al.(2023{\natexlab{a}})Yuan, Chen, Wang, Fung, Peng, and Ji]{yuan2023craft}
Yuan, L., Chen, Y., Wang, X., Fung, Y.~R., Peng, H., and Ji, H.
\newblock Craft: Customizing llms by creating and retrieving from specialized toolsets.
\newblock \emph{arXiv preprint arXiv:2309.17428}, 2023{\natexlab{a}}.

\bibitem[Yuan et~al.(2023{\natexlab{b}})Yuan, Chen, Fu, Ge, Shah, Jankowski, Yang, and Xiao]{Yuan2023DistillingSK}
Yuan, S., Chen, J., Fu, Z., Ge, X., Shah, S., Jankowski, C.~R., Yang, D., and Xiao, Y.
\newblock Distilling script knowledge from large language models for constrained language planning.
\newblock In \emph{Annual Meeting of the Association for Computational Linguistics}, 2023{\natexlab{b}}.
\newblock URL \url{https://api.semanticscholar.org/CorpusID:258564677}.

\bibitem[Zhang et~al.(2024{\natexlab{a}})Zhang, Da, Lee, Robinson, Wu, Song, Zhao, Raja, Slack, Lyu, et~al.]{gsm1k}
Zhang, H., Da, J., Lee, D., Robinson, V., Wu, C., Song, W., Zhao, T., Raja, P., Slack, D., Lyu, Q., et~al.
\newblock A careful examination of large language model performance on grade school arithmetic.
\newblock \emph{arXiv preprint arXiv:2405.00332}, 2024{\natexlab{a}}.

\bibitem[Zhang et~al.(2023{\natexlab{a}})Zhang, Krishna, Awadallah, and Wang]{zhang2023ecoassistant}
Zhang, J., Krishna, R., Awadallah, A.~H., and Wang, C.
\newblock Ecoassistant: Using llm assistant more affordably and accurately.
\newblock \emph{arXiv preprint arXiv:2310.03046}, 2023{\natexlab{a}}.

\bibitem[Zhang et~al.(2023{\natexlab{b}})Zhang, Xia, Wang, Chen, Liu, Wu, and Liu]{zhang2023ideal}
Zhang, S., Xia, X., Wang, Z., Chen, L.-H., Liu, J., Wu, Q., and Liu, T.
\newblock Ideal: Influence-driven selective annotations empower in-context learners in large language models.
\newblock \emph{arXiv preprint arXiv:2310.10873}, 2023{\natexlab{b}}.

\bibitem[Zhang et~al.(2024{\natexlab{b}})Zhang, Zhang, Liu, Song, Wang, Krishna, and Wu]{zhang2024training}
Zhang, S., Zhang, J., Liu, J., Song, L., Wang, C., Krishna, R., and Wu, Q.
\newblock Training language model agents without modifying language models.
\newblock \emph{arXiv preprint arXiv:2402.11359}, 2024{\natexlab{b}}.

\bibitem[Zheng et~al.(2024)Zheng, Gou, Kil, Sun, and Su]{Zheng2024GPT4VisionIA}
Zheng, B., Gou, B., Kil, J., Sun, H., and Su, Y.
\newblock Gpt-4v(ision) is a generalist web agent, if grounded.
\newblock \emph{ArXiv}, abs/2401.01614, 2024.
\newblock URL \url{https://api.semanticscholar.org/CorpusID:266741821}.

\bibitem[Zhou et~al.(2023{\natexlab{a}})Zhou, Xu, Zhu, Zhou, Lo, Sridhar, Cheng, Bisk, Fried, Alon, et~al.]{zhou2023webarena}
Zhou, S., Xu, F.~F., Zhu, H., Zhou, X., Lo, R., Sridhar, A., Cheng, X., Bisk, Y., Fried, D., Alon, U., et~al.
\newblock Webarena: A realistic web environment for building autonomous agents.
\newblock \emph{arXiv preprint arXiv:2307.13854}, 2023{\natexlab{a}}.

\bibitem[Zhou et~al.(2023{\natexlab{b}})Zhou, Jiang, Li, Wu, Wang, Qiu, Zhang, Chen, Wu, Wang, et~al.]{zhou2023agents}
Zhou, W., Jiang, Y.~E., Li, L., Wu, J., Wang, T., Qiu, S., Zhang, J., Chen, J., Wu, R., Wang, S., et~al.
\newblock Agents: An open-source framework for autonomous language agents.
\newblock \emph{arXiv preprint arXiv:2309.07870}, 2023{\natexlab{b}}.

\bibitem[Zhuge et~al.(2023)Zhuge, Liu, Faccio, Ashley, Csord{\'a}s, Gopalakrishnan, Hamdi, Hammoud, Herrmann, Irie, et~al.]{zhuge2023mindstorms}
Zhuge, M., Liu, H., Faccio, F., Ashley, D.~R., Csord{\'a}s, R., Gopalakrishnan, A., Hamdi, A., Hammoud, H. A. A.~K., Herrmann, V., Irie, K., et~al.
\newblock Mindstorms in natural language-based societies of mind.
\newblock \emph{arXiv preprint arXiv:2305.17066}, 2023.

\end{thebibliography}
\bibliographystyle{icml2025}

\newpage
\appendix
\onecolumn

\section{Limitations}
\label{sec:limitation}
The first limitation of our work is cost. A conversation involving the GPT-4 model costs more than a single-agent system. Although we have reduced the cost by decreasing the participant nested group chat agents, it still has a large conversation and profile as context input. The trade-off between performance and cost will become one of the possible future works for further exploration, like window context, conversation pruning, or conversation compression. Another limitation of our work is the lack of thinking about model selection. We should also notice that the current evaluation of LLM is not perfect. Data leaking is widespread in the pertaining process and will cause the misalignment between the test and real-world performance~\citep{gsm1k, xu2024benchmarking}. Therefore, a comprehensive yet fair evaluation is important for us to further discuss the ability of model diversity. In this work, all agents are communicated under AutoGen with a highly customized communication framework. This limited us to generalize \Ours to other existing hand-crafted agents (i.e., absorb those fantastic agents into our agent library). Therefore, a universal agent communication protocol is important.

\section{Difference Between Existing Team-building Frameworks}
\label{apdx:diff}
\begin{table*}[h]
\centering
\caption{\label{tab:cost} Comparison of performance on our reduced dataset for ablation study (see Section~\ref{sec:abla}), where Prog. refers to Programming, DA refers to Data Analysis, Phys. refers to Physics, and Chem. refers to Chemistry. The best results are marked in \best{red bold} and the second best in \secbest{blue}. \Ours achieves the best performance with \texttt{gpt-4-0125-preview} as the backbone. \Ours with \texttt{gpt-4o-mini} can achieve competitive performance with other baselines that use \texttt{gpt-4-0125-preview}, and have significantly lower cost.}
\renewcommand\arraystretch{1.2}
\resizebox{0.7\linewidth}{!}{%
\begin{tabular}{cccccccc}
\hline
\rowcolor[HTML]{B2B2B2} 
  \cellcolor[HTML]{B2B2B2} &
  \cellcolor[HTML]{B2B2B2} &
  \textbf{Math} &
  \textbf{Prog.} &
  \textbf{DA} &
  \textbf{(Sci) Phys.} &
  \textbf{(Sci) Chem.} &
  \textbf{Avg. Rank} \\
\multicolumn{2}{c}{\multirow{-2}{*}{\cellcolor[HTML]{B2B2B2}\textbf{Backbone}}} & \multicolumn{6}{c}{\cellcolor[HTML]{B2B2B2}\textbf{Performance (Accuracy, higher is better)}} \\ \hline
\multicolumn{2}{l}{Vanilla LLM (\texttt{gpt-4-0125-preview})} &
  52.94 &
  72.00 &
  - &
  26.32 &
  31.58 &
  6.8 \\
\multicolumn{2}{l}{Two-Agents (\texttt{gpt-4-0125-preview})} &
  64.71 &
  \secbest{92.00} &
  73.91 &
  47.37 &
  42.11 &
  3.6 \\
\multicolumn{2}{l}{Meta-prompting (\texttt{gpt-4-0125-preview})} &
  70.59 &
  12.00 &
  17.30 &
  \secbest{52.63} &
  52.63 &
  5.0 \\
\multicolumn{2}{l}{AutoAgent (\texttt{gpt-4-0125-preview})} &
  64.71 &
  88.00 &
  52.17 &
  47.37 &
  \best{68.42} &
  3.2 \\
\multicolumn{2}{l}{DyLAN (\texttt{gpt-4-0125-preview})} &
  58.82 &
  \secbest{92.00} &
  - &
  47.37 &
  45.00 &
  - \\
\multicolumn{2}{l}{AgentVerse (\texttt{gpt-4-0125-preview})} &
  64.71 &
  20.00 &
  - &
  36.84 &
  42.11 &
  - \\
\rowcolor[HTML]{E6E6E6} 
\cellcolor[HTML]{E6E6E6} &
  \multicolumn{1}{l}{w/ \texttt{gpt-4-0125-preview}} &
  \best{\best{82.35}} &
  \best{96.00} &
  \secbest{82.60} &
  \best{57.89} &
  \best{68.42} &
  \best{1.2} \\
\rowcolor[HTML]{E6E6E6} 
\cellcolor[HTML]{E6E6E6} &
  \multicolumn{1}{l}{w/ \texttt{gpt-4o-mini}} &
  \secbest{76.47} &
  80.00 &
  \best{91.30} &
  \secbest{52.63} &
  \secbest{57.89} &
  \secbest{2.2} \\
\rowcolor[HTML]{E6E6E6} 
\cellcolor[HTML]{E6E6E6} &
  \multicolumn{1}{l}{w/ \texttt{Llama-3-70B-Instruct}} &
  47.06 &
  80.00 &
  56.52 &
  43.75 &
  36.84 &
  4.6 \\
\rowcolor[HTML]{E6E6E6} 
\multirow{-4}{*}{\cellcolor[HTML]{E6E6E6}\Ours} &
  \multicolumn{1}{l}{w/ \texttt{Llama-3-8B-Instruct}} &
  5.89 &
  48.00 &
  34.78 &
  5.26 &
  5.26 &
  7.4 \\ \hline
\rowcolor[HTML]{B2B2B2} 
\multicolumn{2}{c}{\textbf{Backbone}} & \multicolumn{6}{c}{\cellcolor[HTML]{B2B2B2}\textbf{Cost for Task Completion (US Dollars, lower is better)}} \\ \hline
\multicolumn{2}{l}{Vanilla LLM (\texttt{gpt-4-0125-preview})} &
  1.48 &
  1.08 &
  - &
  \secbest{0.28} &
  1.63 &
  3.8 \\
\multicolumn{2}{l}{Two-Agents (\texttt{gpt-4-0125-preview})} &
  3.10 &
  2.82 &
  5.32 &
  1.34 &
  2.33 &
  5.2 \\
\multicolumn{2}{l}{Meta-prompting (\texttt{gpt-4-0125-preview})} &
  2.92 &
  9.88 &
  8.64 &
  4.18 &
  4.96 &
  5.8 \\
\multicolumn{2}{l}{AutoAgent (\texttt{gpt-4-0125-preview})} &
  4.59 &
  18.32 &
  33.58 &
  12.48 &
  12.28 &
  7 \\
\multicolumn{2}{l}{DyLAN (\texttt{gpt-4-0125-preview})} &
  3.01 &
  8.76 &
  - &
  7.10 &
  8.07 &
  - \\
\multicolumn{2}{l}{AgentVerse (\texttt{gpt-4-0125-preview})} &
  7.63 &
  13.59 &
  - &
  26.34 &
  23.56 &
  - \\
\rowcolor[HTML]{E6E6E6} 
\cellcolor[HTML]{E6E6E6} &
  \multicolumn{1}{l}{w/ \texttt{gpt-4-0125-preview}} &
  7.95 &
  23.67 &
  39.88 &
  15.21 &
  18.68 &
  8 \\
\rowcolor[HTML]{E6E6E6} 
\cellcolor[HTML]{E6E6E6} &
  \multicolumn{1}{l}{w/ \texttt{gpt-4o-mini}} &
  \secbest{0.09} &
  \best{0.03} &
  \secbest{0.29} &
  0.48 &
  \secbest{0.89} &
  \secbest{2} \\
\rowcolor[HTML]{E6E6E6} 
\cellcolor[HTML]{E6E6E6} &
  \multicolumn{1}{l}{w/ \texttt{Llama-3-70B-Instruct}} &
  0.89 &
  1.92 &
  0.89 &
  1.18 &
  1.48 &
  3.4 \\
\rowcolor[HTML]{E6E6E6} 
\multirow{-4}{*}{\cellcolor[HTML]{E6E6E6}\Ours} &
  \multicolumn{1}{l}{w/ \texttt{Llama-3-8B-Instruct}} &
  \best{0.05} &
  \best{0.03} &
  \best{0.02} &
  \best{0.06} &
  \best{0.08} &
  \best{1} \\ \hline
\end{tabular}
}
\end{table*}

In this section, we will discuss the difference between \Ours and other famous agent team-building frameworks, including AutoAgent~\citep{autoagents} AgentVerse~\citep{chen2024agentverse}, and DyLAN~\citep{liu2023dynamic}.

\paragraph{Difference between AgentVerse and \Ours}
Compared with Agentverse, \Ours supports more flexible agent team building and collaboration. AgentVerse includes two types of framework: dynamic team and handcrafted team. The dynamic team completes part of the tasks with the recruitment process, in which some agents are recruited in a fixed process (recruit – chat or comment – evaluate – reflect), and the handcrafted team completes other tasks without the recruitment process. In contrast, we did not design fixed teams for any tasks. Moreover, unlike the fixed sequential process, \Ours can also be involved in the nested group chat as it can solve part of the problems by itself and pass the solution into the nested chat. Furthermore, the \Ours can cache teams in its memory and call back at a proper time. Therefore, the \Ours acts like a time leaper who can participate in different teams on different timelines to help derive better solutions.

\paragraph{Difference between DyLAN and \Ours}
DyLAN is a static build process in which the multi-agent debate starts with a fixed and manually predefined group of experts. On the other hand, DyLAN handcrafts a pool of expert names, their corresponding prompts, and their capabilities. The agent selection from pool to expert group member is manually performed. The framework requires manual effort to function properly.

\section{Case Study Comparing \Ours and Others}
\label{apdx:case}
We demonstrate the benefits and functionalities of \Ours in two distinct scenarios. The first scenario involves the automatic tailoring of specialized agentic systems for solving complex user queries. 
The second scenario highlights its ability to replace manually designed agent systems without human intervention, achieving similar performance to the handcrafted three-agent system OptiGuide~\citep{li2023large} in a specific supply chain optimization scenario.

\begin{figure}[!htb]
\begin{center}
\centerline{\includegraphics[width=1.0\columnwidth]{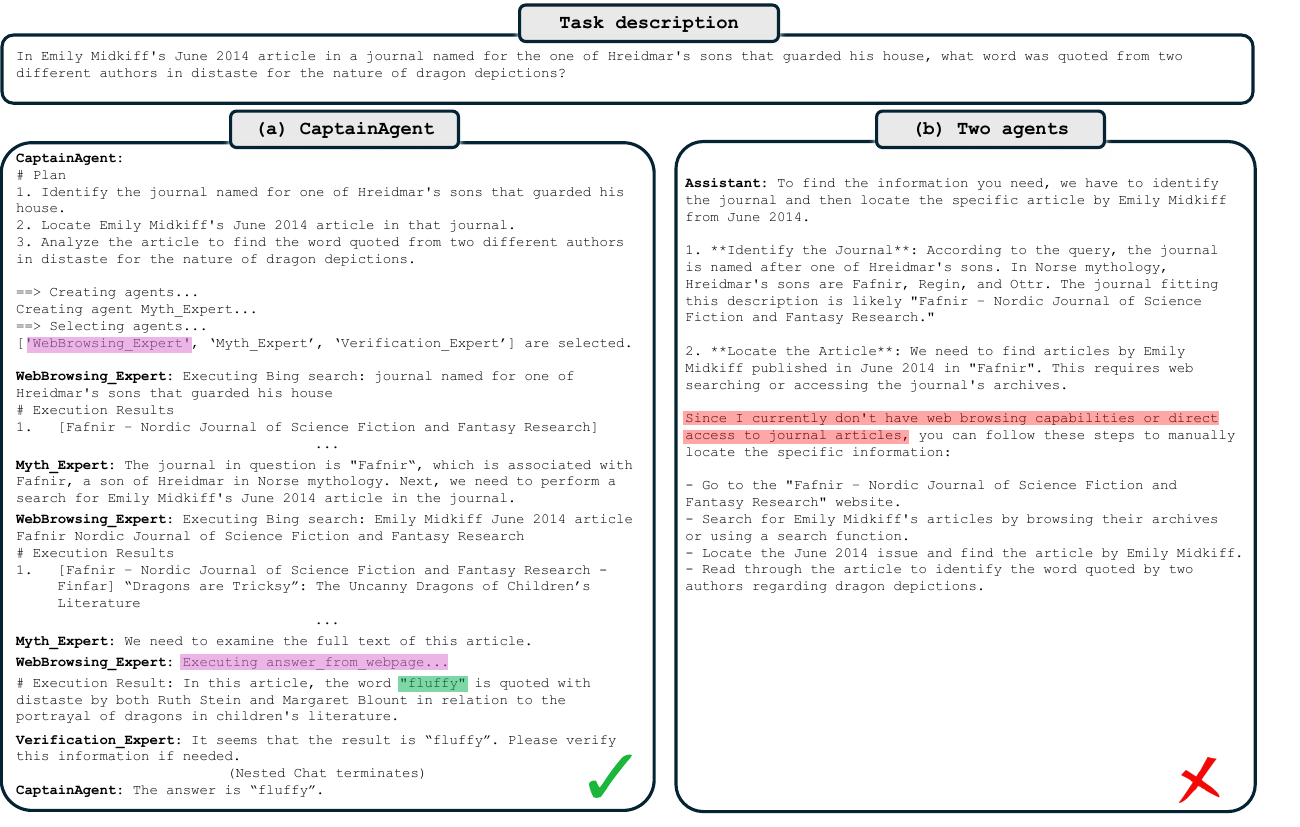}}
\caption{Case study comparing \Ours and Two-agent system from AutoGen. \Ours can aseemble a group of relevant experts to collaboratively handle complex task query from user, while Two-Agent system cannot complete the task due to insufficient capabilities.}
\label{captainagent_case1}
\end{center}
\end{figure}

\begin{figure}[!htb]
\begin{center}
\centerline{\includegraphics[width=1.0\columnwidth]{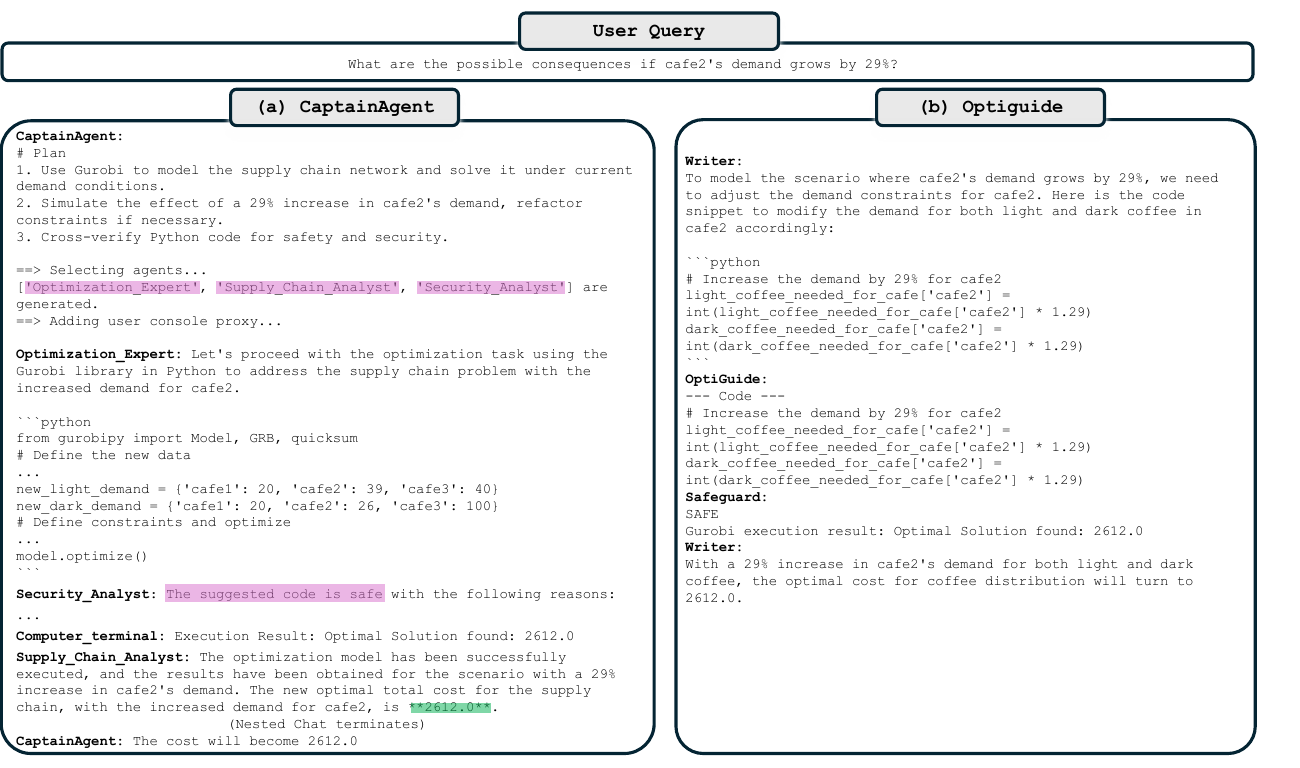}}
\caption{Case study comparing \Ours and Optiguide~\citep{li2023large}, a manually created multi-agent system designed for executing supply chain related query.}
\label{captainagent_case2}
\end{center}
\end{figure}

\subsection{Case Study 1: Arbitrary User Question Answering}
As illustrated in Figure~\ref{captainagent_case1}, a user seeks specific information from an article by Emily Midkiff, published in June 2014. In response, \Ours assembles a team of specialized agents: \texttt{WebBrowsing\_Expert}, \texttt{Myth\_Expert}, and \texttt{Verification\_Expert}. Each agent is designed to address a specific domain, and together they effectively collaborate to solve the problem, under the coordination of \Ours. In contrast, a simple two-agent system from AutoGen fails to resolve the issue due to limitations in the Assistant's capabilities. This highlights \Ours's ability to build an optimal team of agents and ensure their collaboration, whereas basic two-agent systems may lack the necessary functionalities to address more complex queries.

\subsection{Case Study 2: Domain Specific Question Answering}

Figure~\ref{captainagent_case2} illustrates a comparison between \Ours and OptiGuide~\citep{li2023large}, a handcrafted three-agent system designed for supply chain optimization. When provided with the necessary background information, \Ours is able to generate a similar group structure, successfully solving the task. In contrast, an examination of the implementation code reveals that OptiGuide includes a fixed structure requiring over 100 lines of code, meticulously crafted by humans. In comparison, \texttt{CaptainAgent} accomplishes the same task with only a few lines of code, significantly reducing the complexity for developers and easing domain-specific development.

\section{Instruction of \Ours}

We design a general profile message (system message) for \Ours to ensure it can execute our paradigm efficiently and effectively. Instructions are in markdown format, including a planning instruction that can decompose the task into multiple steps, a building instruction (the seek\_experts\_help), a post-seek\_agent\_help instruction, and some general instructions that help task solving.

\subsection{System Message}
\begin{lstlisting}[language={python}]
"""
# Your role
You are a perfect manager of a group of advanced experts.

# How to solve the task
When a task is assigned to you:
1. Analysis of its constraints and conditions for completion. 
2. Response with a specific plan of how to solve the task.

After that, you can solve the task in two ways:
- Delegate the resolution of tasks to other experts created by seeking a group of experts to help and derive conclusive insights from their conversation summarization.
- Analyze and solve the task using your coding and language skills.

# How to seek experts help
The tool "seek_experts_help" can build a group of experts according to the building_task and let them chat with each other in a group chat to solve the execution_task you provided.
- This tool will summarize the essence of the experts' conversation and the derived conclusions.
- You should not modify any task information from meta_user_proxy, including code blocks, but you can provide extra information.
- Within a single response, you are limited to initiating one group of experts.

## building_task
This task helps a build manager to build a group of experts for your task.
You should suggest less than {max_agent_number} roles (including a checker for verification) with the following format.

### Format
- [Detailed description for role 1]
- [Detailed description for role 2]
...
- [Detailed description for verifier]

## execution_task
This is the task that needs the experts to solve by conversation. 
You should Provide the following information in markdown format.

### Format
## Task description
...
## Plan for solving the task
...
## Output format
...
## Constraints and conditions for completion
...
## [Optional] results (including code blocks) and reason from the last response
...

# After seek_experts_help
You will receive a comprehensive conclusion from the conversation, including the task information, results, reason for the results, conversation contradictions or issues, and additional information.
You **must** conduct a thorough verification for the result and reason's logical compliance by leveraging the step-by-step backward reasoning with the same group of experts (with the same group name) when:
- The conversation has contradictions or issues (need double-check marked as yes) or
- The result is different from the previous results.

Note that the previous experts will forget everything after you obtain the response from them. You should provide the results (including code blocks) you collected from the previous experts' responses and put them in the new execution_task.

# Some useful instructions
- You only have one tool called "seek_experts_help."
- Provide a answer yourself after "seek_experts_help".
- You should suggest Python code in a Python coding block (```python...```).
- You must indicate the script type in the code block when using code.
- Do not suggest incomplete code which requires users to modify.
- Be clear about which step uses code, which step uses your language skill, and which step to build a group chat.
- If the code's result indicates an error, fix the error and output the code again. 
- If the error can't be fixed or if the task is not solved even after the code is executed successfully, analyze the problem, revisit your assumption, collect additional info you need, and think of a different approach to try.
- When you find an answer, verify the answer carefully. 
- Include verifiable evidence in your response if possible.
- After completing all tasks and verifications, you should conclude the operation and reply "TERMINATE"
"""
\end{lstlisting}

\subsection{Reflector LLM}
\begin{lstlisting}[language={python}]
"""
# Your task
Briefly summarize the conversation history derived from an experts' group chat by following the answer format.
If you found non-trivial contradictions or issues in the conversation, point it out with a detailed reason and mark the "Need double-check" as "Yes."

# Conversation history:
{chat_history}

# Answer format
## Task
...

## Results
...

## Reason for the results
...

## Contradictions or issues in the conversation
...

### Need to double-check?
[Yes or No]

## Additional information (file path, code blocks, url, etc.)
...
"""
\end{lstlisting}

\subsection{Agent Selector LLM}
\begin{lstlisting}[language={python}]
"""
# Your goal
Match roles in the role set to each expert in the expert set.

# Skill set
{skills}

# Expert pool (formatting with name: description)
{expert_pool}

# Answer format
```json
{{
    "skill_1 description": "expert_name: expert_description", // if there exists an expert that suitable for skill_1
    "skill_2 description": "None", // if there is no experts that suitable for skill_2
    ...
}}
```
"""
\end{lstlisting}

\section{Task Instructions}
\label{apdx:instructions}
We design instructions manually for each scenario and ensure all baselines and \Ours receive the same instructions for a fair comparison\footnote{Except for the information retrieval scenario (GAIA), in which we use the results directly from the leaderboard.}. All instructions include the basic information of the scenario and may suggest some possible Python libraries, including \texttt{pandas}, \texttt{numpy}, \texttt{scipy}, and \texttt{sympy}.

\subsection{Instruction for Mathematics}
\begin{lstlisting}[language={python}]
"""
Please solve the following math problem: 
{problem}
For problems that may be difficult to calculate, try to approximate using Python instead of exact solutions. The following Python packages are pre-installed: sympy, numpy, and scipy. Do not plot any figure.
After verification, reply with the final answer in \\box{{}}.
"""
\end{lstlisting}

\subsection{Instruction for Programming}
\begin{lstlisting}[language={python}]
"""
The following python code imports the `run_tests(candidate)` function from my_tests.py, and runs it on the function `__ENTRY_POINT__`. This will run a set of automated unit tests to verify the correct implementation of `__ENTRY_POINT__`. However, `__ENTRY_POINT__` is only partially implemented in the code below. Complete the implementation of `__ENTRY_POINT__` and output a new stand-alone code block that contains everything needed to run the tests, including: importing `my_tests`, calling `run_tests(__ENTRY_POINT__)`, as well as __ENTRY_POINT__'s complete definition, such that this code block can be run directly in Python.

```python
from my_tests import run_tests

{problem}

# Run the unit tests. All unit tests are running online. DO NOT MODIFY THE FOLLOWING LINE.
run_tests(__ENTRY_POINT__)
```
"""
\end{lstlisting}

\subsection{Instruction for Data Analysis}
\begin{lstlisting}[language={python}]
"""
Let's solve a data analysis problem. Given a CSV file path, you are required to solve a problem following a constraint. Do not plot any figure.

FILE PATH: {file_path}

PROBLEM: {problem}

CONSTRAINT: {constraint}

After verification, reply with the final answer in the format of 
{formats}
"""
\end{lstlisting}

\subsection{Instruction for Science (Chemistry and Physics)}
\begin{lstlisting}[language={python}]
"""
Please solve the following chemistry/physics problem: 
{problem}

Try to approximate using Python instead of using exact solutions for some problems that may be difficult to calculate. The following python packages are pre-installed: sympy numpy scipy. Do not plot any figure.

The required unit of the answer is {unit}.
After verification, reply with the final answer in \\box{{}}.
"""
\end{lstlisting}

\subsection{Instruction for World-information Retreival}
\begin{lstlisting}[language={python}]
"""
# Task
You need to solve the question below given by a user. When you are building tasks, explicitly consider where the task can benefit from web navigation capability.

# Task
{task}
"""
\end{lstlisting}

\section{Agent Library}
\label{apdx:agent-lib}
We provide an example of the agent recorded in the agent library below:
\begin{lstlisting}[language={python}]
{
    "description": "PythonProgramming_Expert is a seasoned authority on rocket physics and classical mechanics, adept in Python programming and utilizing specialized libraries to solve complex aerospace problems with high precision and accuracy.",
    
    "tags": ["gpt-4", "0125", "1106", "claude3", "sonnet", "haiku", 
    
    "gemini-1.5", "llama3", "8b", "70b", "mixtral", "8x22b", "8x7b"],
    
    "name": "PythonProgramming_Expert",
    
    "system_message": "## Your role\nPythonProgramming_Expert is an authoritative specialist in the realm of classical mechanics, with a razor-sharp focus on the intriguing world of rocket physics. This expert boasts a profound understanding of the underlying principles that govern the motion and dynamics of rockets, from their ascent through Earth's atmosphere to their navigation across the vast expanse of space.\n\n## Task and skill instructions\n- Aspiring to the pinnacle of precision and accuracy, PythonProgramming_Expert is armed with an extensive array of numerical methods and approximation techniques. They have mastered the art of formulating and solving complex mathematical problems, using these tools to make precise predictions and optimizations in rocket trajectories and propulsion systems.\n- In addition to their expansive knowledge of physical laws and equations, this expert is a virtuoso in Python programming, wielding libraries like sympy for symbolic mathematics, numpy for numerical computations, and scipy for additional scientific computing capabilities. These tools are the chisels with which PythonProgramming_Expert sculpts solutions to elaborate aerospace quandaries.\n- PythonProgramming_Expert's deft problem-solving abilities are matched only by their meticulous approach to mathematical calculations. Whether confronting a routine calculation or an esoteric formula, they tackle each challenge with the same level of dedication and expertise.\n- Finally, with an unrelenting commitment to veracity, PythonProgramming_Expert rigorously verifies physical and mathematical results. They understand that in the delicate ballet of spaceflight, there is no room for error and that the accurate validation of results is paramount for successful missions. This dedication ensures that when PythonProgramming_Expert presents a solution, it is not only theoretically sound but also practically reliable."
},
\end{lstlisting}

\section{Tool Library}
\label{apdx:tool-lib}
This section provides the names and descriptions of our manually created tool library. The tools are categorized into three classes: Information Retrieval, Data Analysis and Math Problem Solving. For each category, we summarize the patterns of the corresponding dataset and manually craft a set of functions suits the tasks and can potentially enhance the agents' task resolution capability.

\begin{table}[ht]
\centering
\caption{Tools for Information Retrieval category.}
\resizebox{\columnwidth}{!}{
\renewcommand\arraystretch{1.3}
\begin{tabular}{lp{9cm}}
\toprule
\textbf{Tools} & \textbf{Description} \\ \hline
scrape\_wikipedia\_tables & Scrapes Wikipedia tables based on a given URL and header keyword. \\ 
\rowcolor{light-gray} transcribe\_audio\_file & Transcribes the audio file located at the given file path. \\ 
youtube\_download & Downloads a YouTube video and returns the download link. \\
\rowcolor{light-gray} academic\_search & Perform an academic search of papers, authors or an author's papers.\\
docx\_to\_md & Converts a DOCX file to Markdown format. \\
\rowcolor{light-gray}pptx\_to\_md & Convert a PowerPoint presentation (PPTX) to Markdown format. \\
spreadsheet\_to\_md & Convert an Excel spreadsheet file to Markdown format. \\
\rowcolor{light-gray} \multirow{2}{*}{extract\_pdf\_image} & Extracts images from a PDF file and saves them to the specified output directory.\\
extract\_pdf\_text & Extracts text from a specified page or the entire PDF file.\\
\rowcolor{light-gray}get\_youtube\_caption & Retrieves the captions for a YouTube video.\\
image\_qa & Answers your questions about a given image. \\
 \rowcolor{light-gray}optical\_character\_recognition & Perform optical character recognition (OCR) on the given image.\\
 \multirow{4}{*}{perform\_web\_question\_answering }& Perform web search according to keyword and answer your question on each webpage search result, or directly on the webpage if the keyword is a URL. For each search result, a response to the question is provided. \\
\rowcolor{light-gray}scrape\_wikipedia\_tables & Scrapes Wikipedia tables based on a given URL and header keyword. \\
\bottomrule
\end{tabular}
}
\label{tab:tools_ir}
\end{table}

\begin{table}[ht]
\centering
\caption{Tools for Data Analysis category.}
\begin{tabular}{lp{9cm}}
\toprule
\textbf{Tools} & \textbf{Description} \\ \hline
 
calculate\_correlation & Calculate the correlation between two columns in a CSV file. \\
\rowcolor{light-gray} \multirow{2}{*}{calculate\_skewness\_and\_kurtosis} &  Calculate the skewness and kurtosis of a specified column in a CSV file. The kurtosis is calculated using the Fisher definition.\\
 \multirow{2}{*}{detect\_outlier\_iqr} & Detect outliers in a specified column of a CSV file using the IQR method.\\
\rowcolor{light-gray} \multirow{3}{*}{detect\_outlier\_zscore} & Detect outliers in a CSV file based on a specified column. The outliers are determined by calculating the z-score of the data points in the column.\\
 \multirow{2}{*}{explore\_csv} &  Reads a CSV file and prints the column names, shape, data types, and the first few lines of data. \\
\rowcolor{light-gray} \multirow{2}{*}{shapiro\_wilk\_test} & Perform the Shapiro-Wilk test on a specified column of a CSV file.\\
\bottomrule
\end{tabular}
\label{tab:tools_da}
\end{table}

\begin{table}[ht]
\centering
\caption{Tools for Math Problem solving category.}
\begin{tabular}{lp{9cm}}
\toprule
\textbf{Tools} & \textbf{Description} \\ \hline
calculate\_circle\_area\_from\_diameter & Calculate the area of a circle given its diameter.\\
\rowcolor{light-gray} \multirow{2}{*}{ calculate\_day\_of\_the\_week } & Calculates the day of the week after a given number of days starting from a specified day.\\
  \multirow{2}{*}{calculate\_fraction\_sum} & Calculates the sum of two fractions and returns the result as a mixed number.\\
\rowcolor{light-gray}calculate\_matrix\_power & Calculate the power of a given matrix. \\
calculate\_reflected\_point & Calculates the reflection point of a given point about the line y=x.\\
\rowcolor{light-gray} complex\_numbers\_product & Calculates the product of a list of complex numbers.\\
 \multirow{2}{*}{compute\_currency\_conversion} & Compute the currency conversion of the given amount using the provided exchange rate.\\
\rowcolor{light-gray}  \multirow{2}{*}{count\_distinct\_permutations } & Counts the number of distinct permutations of a sequence where items may be indistinguishable.\\
\multirow{2}{*}{evaluate\_expression} & Evaluates a mathematical expression with support for floor function notation and power notation.\\
\rowcolor{light-gray}  \multirow{2}{*}{find\_continuity\_point} & Find the value that ensures the continuity of a piecewise function at a given point.\\
\multirow{2}{*}{fraction\_to\_mixed\_numbers} & Simplifies a fraction to its lowest terms and returns it as a mixed number.\\
\rowcolor{light-gray} \multirow{2}{*}{modular\_inverse\_sum} & Calculates the sum of modular inverses of the given expressions modulo the specified modulus.\\
\multirow{2}{*}{simplify\_mixed\_numbers } & Simplifies the sum of two mixed numbers and returns the result as a string in the format 'a b/c'.\\
\rowcolor{light-gray}sum\_of\_digit\_factorials & Calculates the sum of the factorial of each digit in a number.\\
sum\_of\_primes\_below & Calculates the sum of all prime numbers below a given threshold.\\

\bottomrule
\end{tabular}
\label{tab:tools_math}
\end{table}

\end{document}